\documentclass[lettersize,journal]{IEEEtran}
\usepackage{amsmath,amsfonts}
\usepackage{algorithmic}
\usepackage{algorithm}
\usepackage{array}
\usepackage{textcomp}
\usepackage{stfloats}
\usepackage{url}
\usepackage{verbatim}
\usepackage{graphicx}
\usepackage{multirow}
\usepackage{float}
\usepackage{graphicx}
\usepackage{pifont} 
\usepackage{booktabs} 
\usepackage{placeins}
\usepackage{color}
\usepackage{bm}
\usepackage{enumitem}
\newcommand{\cmark}{\ding{51}}

\def\BibTeX{{\rm B\kern-.05em{\sc i\kern-.025em b}\kern-.08em
    T\kern-.1667em\lower.7ex\hbox{E}\kern-.125emX}}
\usepackage{balance}
\usepackage[caption=false,font=normalsize,labelfont=sf,textfont=sf]{subfig}
\usepackage{subcaption}
\usepackage{graphicx}
\usepackage{subfig}

\begin{document}
\title{AdaMuS: Adaptive Multi-view Sparsity Learning for Dimensionally Unbalanced Data}
\author{Cai~Xu,
        Changhao~Sun,
        Ziyu~Guan,
        and~Wei~Zhao%

\thanks{C. Xu, C. Sun, Z. Guan, and W. Zhao are with Xidian University, Xi'an 710071, China. Specifically, C. Sun is with the School of Cyber Engineering. C. Xu, Z. Guan, and W. Zhao are with the School of Computer Science and Technology (e-mail: cxu@xidian.edu.cn; changhaosun@stu.xidian.edu.cn; zyguan@xidian.edu.cn; ywzhao@xidian.edu.cn).}%
\thanks{(Corresponding author:Wei Zhao .)}%
}
\markboth{IEEE Transactions on Image Processing,~Vol.~XX, No.~XX, January~2026}%
{Xu \MakeLowercase{\textit{et al.}}: AdaMuS: Adaptive Multi-view Sparsity Learning}

\maketitle


\begin{abstract}
Multi-view learning primarily aims to fuse multiple features to describe data comprehensively.
Most prior studies implicitly assume that  different views share similar dimensions. In practice, however, severe dimensional disparities often exist among different views, leading to the unbalanced multi-view learning issue. For example, in emotion recognition tasks, video frames often reach dimensions of $10^6$, while physiological signals comprise only $10^1$ dimensions. Existing methods typically face two main challenges for this problem: (1) They often bias towards high-dimensional data, overlooking the low-dimensional views. (2) They struggle to effectively align representations under extreme dimensional imbalance, which introduces severe redundancy into the low-dimensional ones.
To address these issues, we propose the Adaptive Multi-view Sparsity Learning (AdaMuS) framework. First, to prevent ignoring the information of low-dimensional views, we construct view-specific encoders to map them into a unified dimensional space. Given that mapping low-dimensional data to a high-dimensional space often causes severe overfitting, we design a parameter-free pruning method to adaptively remove redundant parameters in the encoders. Furthermore, we propose a sparse fusion paradigm that flexibly suppresses redundant dimensions and effectively aligns each view. Additionally, to learn representations with stronger generalization, we propose a self-supervised  learning paradigm that obtains supervision information by constructing similarity graphs. Extensive evaluations on a synthetic toy dataset and seven real-world benchmarks demonstrate that AdaMuS consistently achieves superior performance and exhibits strong generalization across both classification and semantic segmentation tasks. The code is released at \url{https://github.com/Hayd-coder/AdaMuS}.

\end{abstract}


\section{Introduction}
\IEEEPARstart{M}{ulti-view} data is prevalent in real-world scenarios. For example, emotion recognition comprehensively integrates subjects' video frames and physiological signals. These different modals provide consistent and complementary properties for the same data instance. Fusing multi-modal data allows for a more comprehensive description, thereby boosting the performance of various downstream tasks, such as classification \cite{li2025community}, clustering \cite{liu2025two,long2025tlrlf4mvc}, retrieval \cite{ventura2024covr}, recognition \cite{yang2024cross}, and segmentation \cite{sun2024uni}. With the
rapid advancements in deep learning, deep multi-view learning has recently seen
massive progress \cite{ngiam2011multimodal}. This paper focuses on its fundamental problem:
Multi-view Representation Learning (MRL), which aims to extract a comprehensive
representation containing the consistent and complementary information across
all views.






\begin{figure*}[t]
    \centering
    \includegraphics[width=\textwidth]{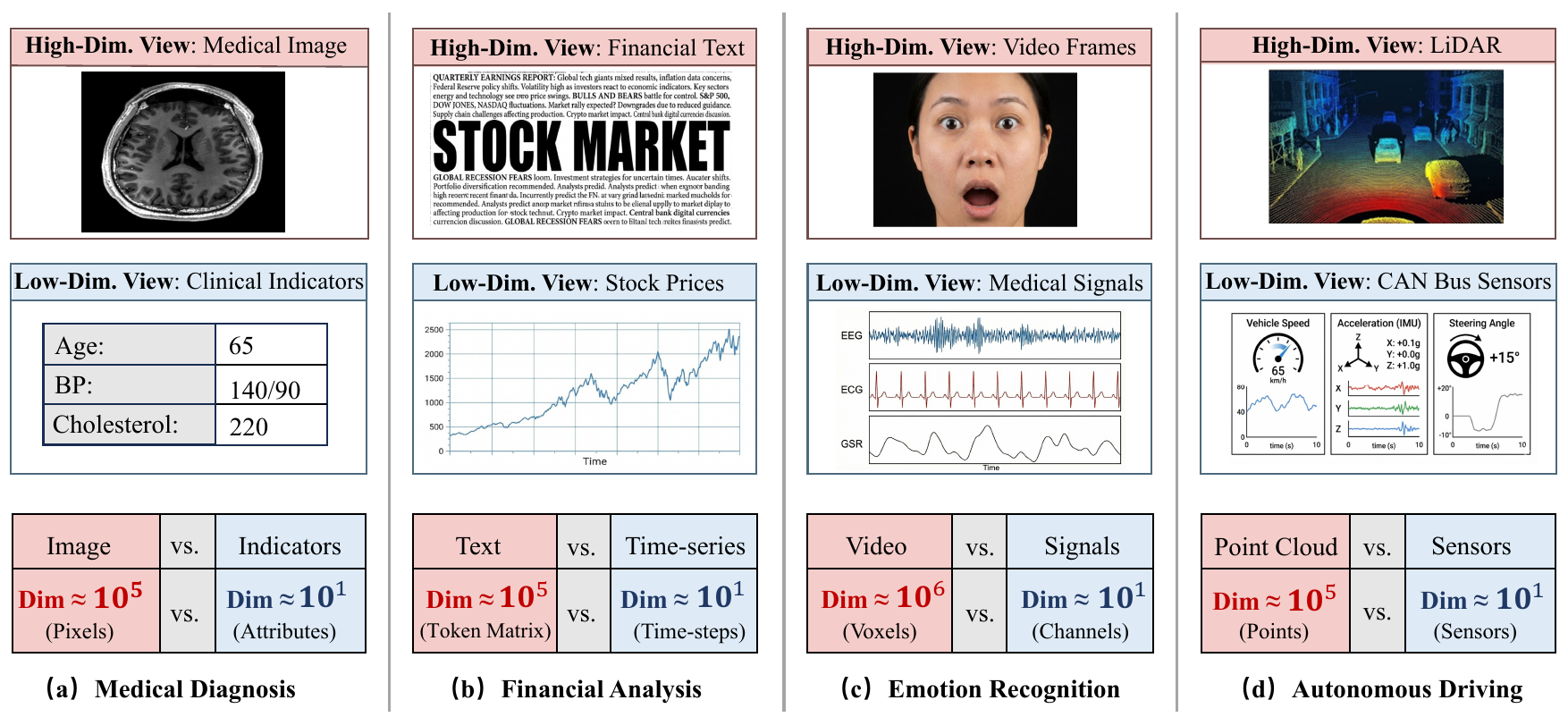}
    \caption{\textbf{Ubiquitous Dimensional Imbalance in Real-world Multi-view Applications.} 
    We illustrate four representative scenarios—(a) Medical Diagnosis, (b) Financial Analysis, (c) Emotion Recognition, and (d) Monitoring—where extreme dimensionality disparities (e.g., $\approx 10^6$ vs. $\approx 10^1$) between views are prevalent. This fundamental imbalance challenges conventional fusion methods, which often bias towards high-dimensional features.}
    \label{motivation}
\end{figure*}

Existing MRL frameworks \cite{huang2021deep,wen2025partial,ke2024rethinking}  implicitly assume that different views have similar dimensions. However, real-world data often violates this assumption. For example, in emotion recognition tasks, both video frames and physiological signals (such as EEG and ECG) are important. Video data often has dimensions exceeding \(10^6\), while physiological signals typically have dimensions on the order of \(10^1\) \cite{koelstra2011deap}. Furthermore, in medical diagnosis, CT images and blood test metrics (such as white blood cell count and hemoglobin levels), which reflect patients' physiological health indicators are also  highly unbalanced \cite{kline2022multimodal}, giving rise to the Unbalanced Multi-view Representation Learning problem \cite{xu2023unbalanced}.
Existing methods typically suffer from the following drawbacks for this problem: (1) They often overlook the information of low-dimensional views. For example, directly concatenating high-dimensional and low-dimensional data  \cite{baltruvsaitis2018multimodal,zhao2024deep} often results in the low-dimensional information being overwhelmed by the high-dimensional information. This causes the model to degenerate into single-view learning dominated by high-dimensional features  \cite{wang2020makes,peng2022balanced}, failing to truly leverage the complementary  multi-view information. (2) They struggle
to effectively align features under extreme dimensional imbalance, which introduces severe redundancy into the low-dimensional ones. Specifically, existing methods first using DNNs to construct balanced view-specific representations and then perform fusion \cite{andrew2013deep,xu2023aaai,zhang2020deep}. However, this blind alignment leads to two consequences: First, for low-dimensional views, directly introducing DNNs to expand dimensions brings a large number of redundant parameters. The mismatch between complex networks and limited information can easily lead to overfitting \cite{zhao2024deep}. Second, the forcibly expanded dimensions often lack practical meaning. Once these redundant dimensions are fused into the final comprehensive representation, they severely compromise its reliability \cite{xu2023unbalanced}.



\begin{figure*}
    \centering
    \includegraphics[width=\linewidth]{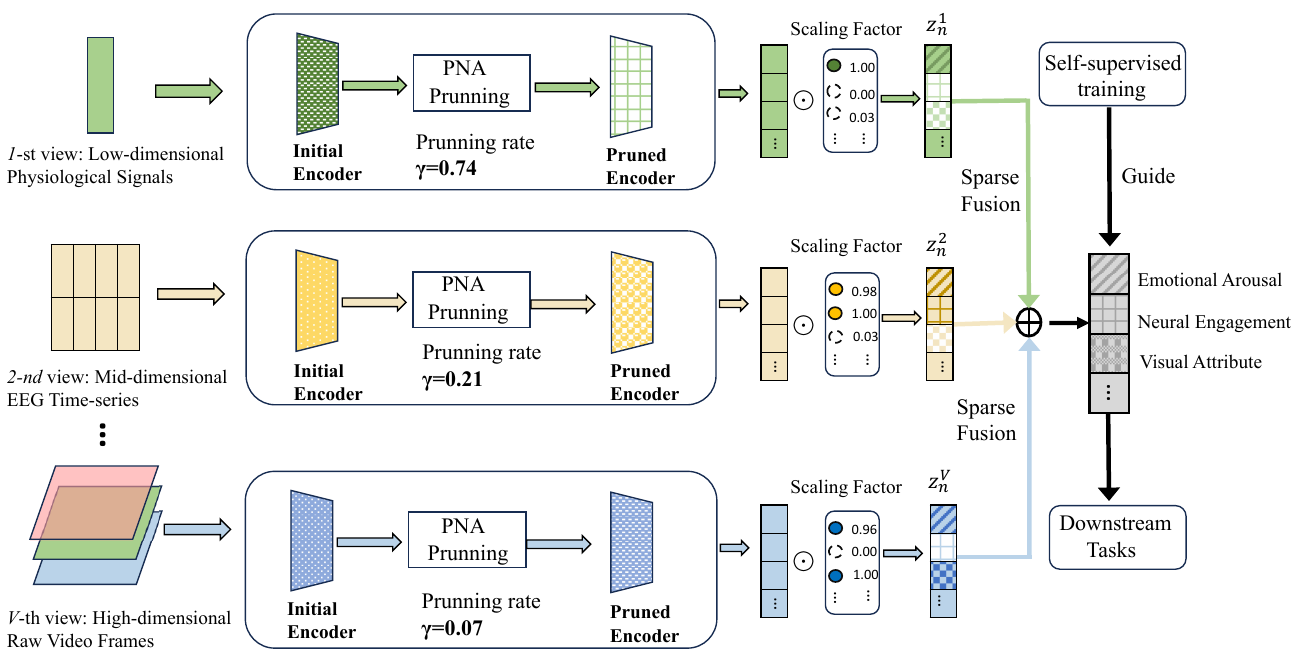}
    \caption{\label{fig:framework} Illustration of the proposed AdaMuS framework. AdaMuS establishes view-specific encoders $f^{(v)}(\cdot)$ to process unbalanced multi-view data. These encoders are structurally optimized by the \textbf{Principal Neuron Analysis (PNA)} module to learn aligned representations $\{z_n^v\}_{v=1}^V$ with a unified dimension. Specifically, a \textbf{Multi-view Sparse Batch Normalization (MSBN)} layer is proposed to explicitly integrate these features via a sparse fusion paradigm. Finally, we train AdaMuS in a self-supervised manner to guide the representation learning.}
\end{figure*}

To address the above limitations, we propose a new framework called Adaptive Multi-view Sparsity Learning for Dimensionally Unbalanced Data (AdaMus). As illustrated in Figure~\ref{fig:framework}, we first construct view-specific encoders to learn aligned representations. To avoid the overfitting problem caused by expanding low-dimensional views, we design a parameter-free pruning method called Principal Neuron Analysis (PNA). PNA computes a unique pruning rate for each view by jointly analyzing neuron correlations and the imbalance degree of the specific view. Based on the pruning rate, redundant neurons are removed from each view-specific encoder, so as to mitigate the severe overfitting risk caused by forced dimensional expansion. Second, considering that some dimensions in the view-specific representations are irrelevant to the final fused representation, we design a Multi-view Sparse Batch Normalization (MSBN) layer. By adding an $l_1$-norm constraint to the loss function, this layer penalizes the scaling factors of redundant dimensions. Finally, we propose a self-supervised contrastive learning paradigm guided by balanced view-specific similarity graphs, thereby extracting generalizable representations for diverse downstream tasks.

The key contributions of our paper are listed below:

(1) We propose a novel deep learning framework named AdaMuS, specifically designed for problem of Unbalanced Multi-view Representation Learning. It effectively prevents overlooking low-dimensional information and eliminates the severe redundancy introduced by forced dimensional alignment.

(2) We design a parameter-free Principal Neuron Analysis (PNA) method to adaptively prune redundant parameters, effectively mitigating the severe overfitting caused by expanding low-dimensional views.

(3) We develop a Multi-view Sparse Batch Normalization (MSBN) layer to achieve high-quality cross-view alignment and suppress redundant dimensions generated by forced alignment. Furthermore, we propose a self-supervised contrastive learning paradigm that derives supervision signals from balanced similarity graphs, yielding generalizable representations for diverse downstream tasks.

(4) Extensive experiments on a toy example and seven real-world benchmarks demonstrate that AdaMuS consistently achieves superior performance and exhibits strong generalization capabilities in both classification and semantic segmentation tasks.

\section{Related Work}
\subsection{Deep MCRL}
Deep Multi-view Comprehensive Representation Learning (MCRL) employs Deep Neural Networks (DNNs) to extract shared and complementary semantics from multiple views. These approaches generally fall into two categories: Multi-view Synergistic Representation Learning (MSRL) and Multi-view Aligned Representation Learning (MARL).

MSRL utilizes distinct encoders $\{f_s^v\}_{v=1}^V$ to map inputs $\{\boldsymbol{x}^v\}_{v=1}^V$ into view-specific embeddings $\{\boldsymbol{z}_s^v\}_{v=1}^V$. These embeddings are combined into a synergistic representation $\boldsymbol{z}_s$ via an aggregation function $f_s(\cdot)$ defined as:
\begin{equation} \label{eq:1}
    \boldsymbol{z}_s = f_s(\boldsymbol{z}_s^1, \dots, \boldsymbol{z}_s^V), \quad \boldsymbol{z}_s^v = f_s^v(\boldsymbol{x}^v),
\end{equation}
where $f_s(\cdot)$ typically involves concatenation or logical operations \cite{ngiam2011multimodal,arevalo2017gated}. MSRL optimization often targets reconstruction accuracy \cite{wan2021cross, he2022masked} or correlation maximization \cite{andrew2013deep}. However, applying standard MSRL to unbalanced data is problematic. Enforcing balanced embedding dimensions causes information loss in high-dimensional views and creates redundancy in low-dimensional ones.

Alternatively, MARL approaches \cite{nie2016parameter, xu2024reliable,liu2024robust,wang2025seqmvrl} enforce strict alignment between view-specific features $\{\boldsymbol{z}_a^v\}_{v=1}^V$. The unified representation is derived through a weighted average:
\begin{equation} \label{eq:2}
    \boldsymbol{z}_a = \sum_{v=1}^V \omega^v \boldsymbol{z}_a^v, \quad \boldsymbol{z}_a^v = f_a^v(\boldsymbol{x}^v),
\end{equation}
where $\omega^v$ represents the scalar weight for the $v$-th view and $f_a^v(\cdot)$ is the specific encoder.

A critical limitation of MARL lies in its fusion mechanism. It assigns a global weight $\omega^v$ to all dimensions of a view. This strategy fails when dimensional discrepancies are severe. High-dimensional views contain semantic details absent in low-dimensional ones. Feedback from the fused representation forces the low-dimensional encoders to generate ``pseudo-dimensions'' for alignment. These artificial features obscure genuine intrinsic information. Consequently, the model treats alignment noise as valid signals. This process erodes the uniqueness of the smaller view, resulting in a model dominated by the high-dimensional input. To rectify this, our AdaMuS framework introduces the Multi-view Sparse Batch Normalization (MSBN) layer. MSBN models the individual contribution of each latent dimension. It enables the model to selectively suppress redundant dimensions, ensuring balanced and effective fusion.

\subsection{Unbalanced Multiview Learning}
In real-world multiview data, the dimensionalities of different views often vary considerably, leading to the problem of unbalanced multiview learning. Existing solutions can be broadly divided into three categories.

A straightforward solution employs deep neural networks (DNNs) to project heterogeneous views into a unified representation space, directly mitigating dimensional imbalance~\cite{yu2024deep,xu2021adversarial}. However, Xu et al.~\cite{xu2023unbalanced} noted that such mappings can severely overfit low-dimensional views due to excessive network parameters.

Another approach adopts decision-level fusion, where each view is processed by an independently trained network and the resulting predictions are fused. For instance, Liu et al.~\cite{liu2018late} cluster each view separately before aggregating the results, whereas Han et al.~\cite{han2021trusted} train a ``trust decision'' network for each view to estimate class-wise evidence, assigning greater weights to more reliable views in the final fusion stage.
A third category constructs balanced similarity graphs for each view, regardless of their original feature dimensionalities. The constructed graphs are then fused to obtain the final feature representation~\cite{zhang2016flexible}. Although decision-level fusion and similarity-graph methods avoid direct feature-level fusion, they keep the views separate until predictions or latent representations are produced. This limits their ability to capture correlations between views. Moreover, decision-level fusion is tailored to category-oriented objectives such as classification and clustering, which limits its transfer to downstream tasks like segmentation.

To mitigate the overfitting caused by DNNs in low-dimensional views, UMDL~\cite{xu2023unbalanced} builds an overcomplete dictionary where the number of atoms is larger than the original dimension. It then maps low-dimensional samples into a higher-dimensional space, producing representations that amplify structural differences and enable DNNs to learn more discriminative features. However, this approach requires manually specifying which views are ``low-dimensional.'' In practice, the boundary between high and low dimensionality is often ambiguous, and such subjective partitioning can be inaccurate, which in turn constrains the method’s applicability.

Our AdaMuS is a representation-level fusion framework and can be applied to a wide range of downstream tasks, including segmentation. It avoids manual partitioning of high- and low-dimensional views, while integrating a parameter-free pruning strategy to mitigate overfitting in DNNs.

\subsection{Network Pruning}
Network pruning is a powerful method to prevent overfitting by eliminating unnecessary parameters in neural networks \cite{blalock2020state,cheng2017survey}. Generally, this technique includes two types: unstructured pruning \cite{han2015learning}, which deletes specific weights, and structured pruning \cite{li2016pruning,liu2017learning,luo2017thinet}, which discards whole neurons or filters. Structured pruning has become a research hotspot due to its amenability to hardware acceleration, and our work focuses on this area.

Mainstream structured pruning strategies assess neuron importance using various criteria, such as examining the magnitude of a neuron's activations on a dataset~\cite{luo2017thinet}, using gradient information to estimate the impact on model loss upon its removal~\cite{molchanov2016pruning}, or introducing sparsity constraints during training to automatically drive the weights of irrelevant neurons toward zero~\cite{liu2017learning}.

A limitation of existing pruning methods is their reliance on a manually-set global pruning rate \cite{li2016pruning,liu2017learning,he2018soft}, a drawback particularly pronounced in unbalanced multi-view learning. Given that the dimensionality and information redundancy can vary dramatically across views, a fixed pruning rate may result in "under-pruning" high-dimensional views while "over-pruning" low-dimensional ones, thereby impairing final performance \cite{liu2019metapruning,luo2020autopruner}. To overcome this, we propose a parameter-free, adaptive pruning strategy that automatically computes an optimal pruning rate for each view-specific encoder based on its intrinsic data characteristics.

\section{Method}
\subsection{Problem Definition}
We consider a multi-view dataset $\mathcal{X} = \{X_n\}_{n=1}^N$ comprising $N$ instances. Each instance $X_n$ contains a set of feature vectors $\{\boldsymbol{x}_n^v\}_{v=1}^V$, where $\boldsymbol{x}_n^v \in \mathbb{R}^{D_v}$ corresponds to the $v$-th view. A critical characteristic of real-world scenarios is that the feature dimensionality $D_v$ varies drastically across views. To formally quantify this dimensional disparity, we introduce a metric termed the Unbalanced Degree, denoted as $\Lambda$:
\begin{equation} \label{eq:3}
    \Lambda = \underbrace{\frac{1}{C_V^2} \sum_{v_i=1}^V \sum_{\substack{v_j=1 \\ v_j \neq v_i}}^V \frac{2|D_{v_i} - D_{v_j}|}{D_{v_i} + D_{v_j}}}_{\text{Pairwise Component}} + \underbrace{\frac{1}{\bar{D}} \sqrt{\frac{1}{V} \sum_{v = 1}^{V} (D_v - \bar{D})^2}}_{\text{Global Component}},
\end{equation}
where $\bar{D} = \frac{1}{V} \sum_{v = 1}^{V} D_v$ represents the mean dimensionality, and $C_V^2 = \frac{V(V - 1)}{2}$ counts the total view pairs.

The metric consists of two normalized parts to allow fair comparisons across different datasets. The pairwise term evaluates local disparities by averaging the dimensional differences between view pairs. Meanwhile, the global term assesses the overall dataset imbalance through the variance of all view dimensions. A larger $\Lambda$ represents a higher degree of imbalance, and $\Lambda=0$ indicates that all views are perfectly balanced with identical dimensions.

The primary aim is to construct a multi-view comprehensive representation $z_n \in \mathbb{R}^H$ that combines the consistent and complementary information derived from the multi-view data $\{x_n^v\}_{v=1}^V$, despite their diverse unbalance degrees $\Lambda$.

\subsection{The AdaMuS Framework}
As shown in Fig.~\ref{fig:framework}, the AdaMuS architecture consists of two core parts: adaptive multi-view sparse network learning and adaptive cross-view sparse alignment learning.

First, the adaptive multi-view sparse network learning module constructs view-specific encoders to learn aligned representations $\{z_{n}^{v}\}_{v=1}^{V}$. To avoid the overfitting problem caused by expanding the dimensions of low-dimensional views, we design a parameter-free pruning method called Principal Neuron Analysis (PNA). This method automatically removes redundant neurons from each view-specific encoder. Inspired by the Principal Component Analysis (PCA) algorithm, PNA computes a unique pruning rate for each view by jointly analyzing the correlations of the neurons and the unbalanced degree of the specific view.
Second, the adaptive cross-view sparse alignment learning module handles the feature fusion. Considering that some dimensions in the view-specific representations $\{z_{n}^{v}\}_{v=1}^{V}$ are irrelevant to the final fused representation $z_n$, we design a Multi-view Sparse Batch Normalization (MSBN) layer. This layer eliminates the negative impact of redundant dimensions. 
Finally, we design a loss function to drive the forward learning process. We train the model using a self-supervised contrastive learning method. The supervision information is constructed from balanced view-specific similarity graphs.

\subsubsection{Adaptive Multi-view Sparse Network Learning }
In this module, we first use view-specific encoders (DNNs) to project the original features into an aligned space to avoid ignoring the low-dimensional views. However, expanding low-dimensional data to a high-dimensional space may cause severe overfitting. To mitigate this issue, we propose an adaptive pruning method called Principal Neuron Analysis (PNA).  This method automatically removes redundant parameters, transforming the dense encoders into sparse networks that match the actual information density of each view.

Specifically, to avoid ignoring the information of low-dimensional views, AdaMuS first establishes view-specific encoders $\{f^v(\cdot; \boldsymbol{\Theta}_f^v)\}_{v=1}^V$ to map the original features of the $n$-th sample $\boldsymbol{x}_n^v$ to an aligned view-specific representation $\boldsymbol{z}_n^v \in \mathbb{R}^H$ in a unified dimension space as follows:
\begin{equation} \label{eq:4}
    \boldsymbol{z}_n^v = f^v(\boldsymbol{x}_n^v; \boldsymbol{\Theta}_f^v),
\end{equation}
where $\boldsymbol{x}_n^v$ is the input feature vector for the $v$-th view, and $\boldsymbol{\Theta}_f^v$ represents the set of trainable parameters for the corresponding encoder.

In unbalanced multi-view learning, applying Deep Neural Networks (DNNs) to low-dimensional views often results in a severe mismatch between model complexity and data information. The excessive parameters relative to the low input dimensions typically lead to overfitting \cite{han2016dsd,cheng2024survey}. 
\begin{figure}[t]
\centering
\includegraphics[width=2.1in]{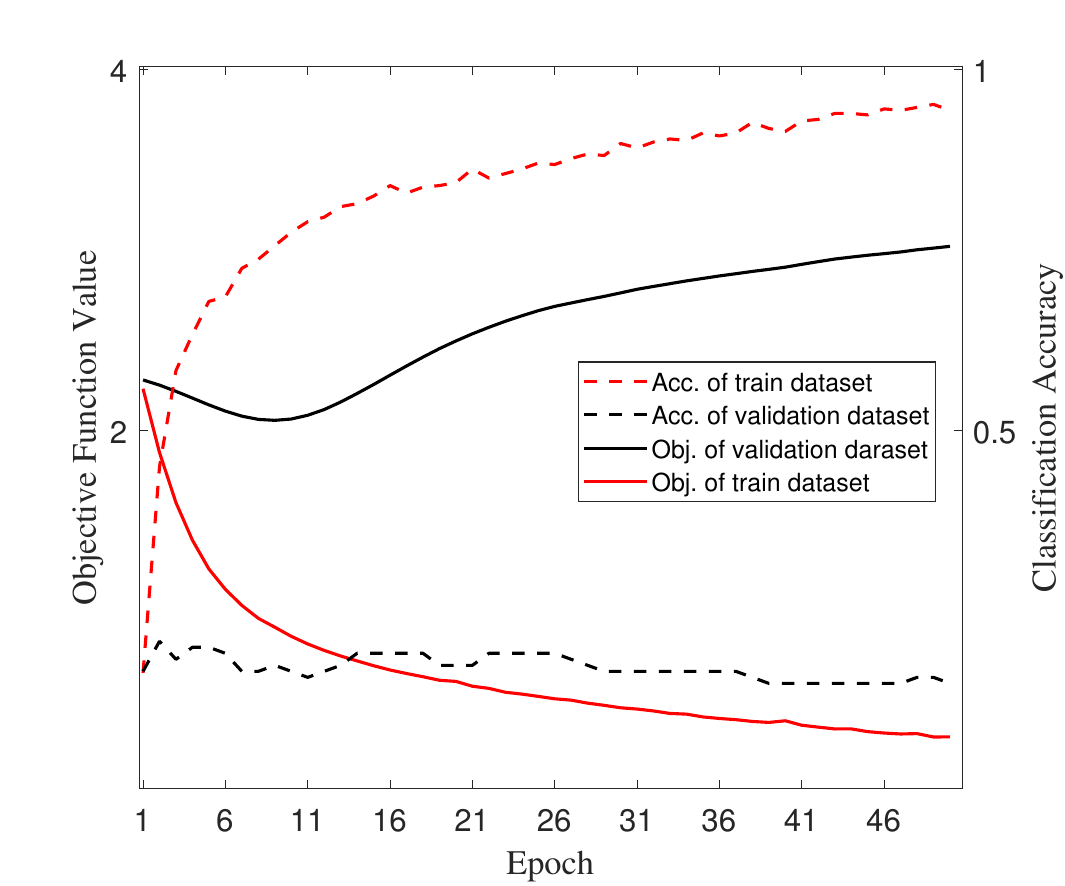}
\caption{Learning curves of a DNN classifier on the low-dimensional view of the CUB dataset. The model projects the 30-dimensional input into a high-dimensional space (structure: 30–1024–512–10). The increasing gap between training and validation performance demonstrates that expanding low-dimensional data with deep networks leads to overfitting.}
\label{DNN-overfitting}
\end{figure}

We demonstrate this phenomenon using the CUB dataset, which contains two views with significantly different dimensions (30 vs. 1024). We trained a single-view DNN to project the 30-dimensional view into the 1024-dimensional space. The results in Fig.~\ref{DNN-overfitting} show that while the training error decreases steadily, the validation error rises. This suggests that directly applying dense DNNs for dimensional expansion significantly increases the risk of overfitting.

To address this, we introduce network pruning into the multi-view learning framework to reduce parameter redundancy in the encoders $\boldsymbol{\Theta}_f^{v}$. Specifically, we employ structured pruning to remove redundant neurons and their associated connections layer by layer. This process results in a compressed parameter set $\boldsymbol{\Theta}_f^{v'}$, which is then used to compute the aligned representation $\boldsymbol{z}_n^v$.
\begin{equation} \label{eq:15}
    \boldsymbol{z}_n^v = f^{v'}(\boldsymbol{*}; \boldsymbol{\Theta}_f^{v'}).
\end{equation}

Traditional pruning methods~\cite{luo2017thinet, molchanov2016pruning,liu2017learning}  rely on a global pruning rate, a manually preset hyper-parameter based on model performance. In the unbalanced multi-view learning, this approach becomes inconvenient as the number of encoders increases. Furthermore, due to the great differences in dimensionality and information redundancy among views, a fixed rate can lead to the ``under-pruning" of high-dimensional views and the ``over-pruning" of low-dimensional ones. 

In this work, we propose a new parameter-free pruning
method, PNA, to remove neurons in each layer, that have
high correlations with the others, and retrain the rest. Unlike those manual methods, PNA can automatically calculates the specific pruning rate for every layer based on the intrinsic redundancy of the view's encoder.

Inspired by PCA algorithm, PNA determines the pruning rate for the $l$-th layer based on the correlation within its neurons' output, $\boldsymbol{Z}_l^v = [\boldsymbol{z}_{1,l}^v, \cdots, \boldsymbol{z}_{N,l}^v]^T \in \mathbb{R}^{N \times D_l^v}$. Our premise is that a higher correlation between neurons should correspond to a higher pruning rate. Pruning neurons in each layer is essentially equivalent to dimensionality reduction, which reduces the processed data dimension, $k$, to be less than $D_l^v$. 

PCA operates by computing eigenvalues and eigenvectors from the covariance matrix. It then reconstructs the data using only the eigenvectors associated with the top-$k$ largest eigenvalues. In special cases, if the eigenvalues follow a uniform distribution, it implies that the features of the original data are uncorrelated, making dimensionality reduction unnecessary. Conversely, if the eigenvalues follow the Dirac $\delta$-distribution, it indicates that the features are highly correlated.





\begin{figure}[t]
    \centering
    \includegraphics[width=\linewidth]{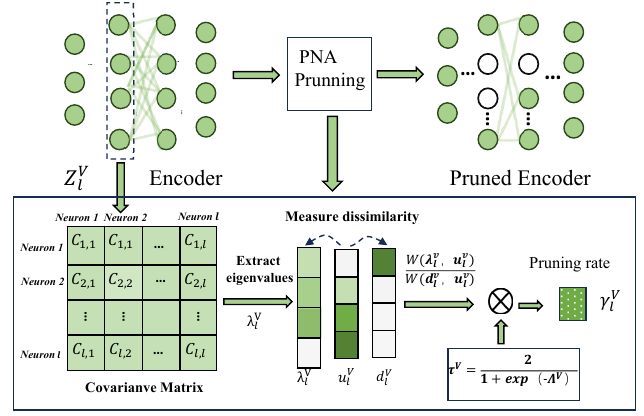}
    \caption{\label{fig:pna_process} PNA workflow: (1) Compute covariance matrix from layer outputs; (2) Extract eigenvalue distribution; (3) Measure dissimilarity to Uniform/Dirac baselines; (4) Adjust pruning rate with view imbalance.}
\end{figure}

Based on the above analysis, we compute the covariance matrix of $\boldsymbol{Z}_l^v$ and extract its eigenvalues, denoted as $\boldsymbol{\lambda}_l^v \in \mathbb{R}^{D_l^v}$, which are then normalized to sum to one. We deem that $\boldsymbol{\lambda}_l^v$ can provide insight into the correlation of neurons' output. The discrepancy between $\boldsymbol{\lambda}_l^v$ and two theoretical distributions quantifies the level of redundancy: the closer $\boldsymbol{\lambda}_l^v$ is to a uniform distribution $\boldsymbol{u}_l^v \in \mathbb{R}^{D_l^v} \sim U[1, D_l^v]$, the less correlated the neurons' outputs are, suggesting that most neurons should be preserved. In contrast, if $\boldsymbol{\lambda}_l^v$ closely resembles the Dirac $\delta$-distribution $\boldsymbol{d}_l^v$, it indicates that the neuron outputs are highly correlated. Consequently, mass neurons should be pruned to remove this redundancy.

Next, to quantify this relationship, we measure the dissimilarity between the empirical eigenvalue distribution $\boldsymbol{\lambda}_l^v$ and the ideal uncorrelated case $\boldsymbol{u}_l^v$ using the Wasserstein metric $W(\boldsymbol{\lambda}_l^v, \boldsymbol{u}_l^v)$. This metric is bounded within the range $[0, W(\boldsymbol{d}_l^v, \boldsymbol{u}_l^v)]$. We then define the pruning rate $\gamma_l^v$ as the normalized dissimilarity:
\begin{equation} \label{eq:15}
    \gamma_l^v = \frac{W(\boldsymbol{\lambda}_l^v, \boldsymbol{u}_l^v)}{W(\boldsymbol{d}_l^v, \boldsymbol{u}_l^v)} \tau^v.
\end{equation}

To prevent reducing the expressive power of DNNs caused by the high pruning rate, we introduce a moderating factor $\tau^v \in [0, 1]$ to control the final pruning rate $\gamma_l$. We hypothesize that the unbalanced degree of each view, denoted as $\Lambda^v$, is inversely proportional to the capacity it should retain. This relationship is formulated as follows:
\begin{align}
    \tau^v &= \frac{2}{1 + \exp(-\Lambda^v)}, \label{eq:16} \\
    \Lambda^v &= \frac{|D_v - \bar{D}|}{2\sigma_D} + \frac{|p^v - \bar{p}|}{2\sigma_p}, \label{eq:17}
\end{align}
where $p^v$ represents the proportional relationship between the dimensions of $\boldsymbol{z}_n^v$ and $D_v$. The terms $\bar{D}$ ($\bar{p}$) and $\sigma_D$ ($\sigma_p$) are the mean and variance of $D$ ($p$), respectively.

Once the final pruning rate is determined, we proceed to remove the most correlated neurons in each layer. For each neuron, we first compute the $L_1$-norm of its Pearson's correlation coefficients with all other neurons in the same layer. Subsequently, we remove the top $\gamma_l \times D_l^v$ neurons that exhibit the maximum norm. It is important to note two key aspects of our pruning stage: (1) The initial model is first pre-trained according to Eq. (13) before the encoders of all views are pruned. (2) We employ a one-shot pruning strategy, which removes all targeted neurons from all layers simultaneously. Consequently, the pruning rates $\gamma_l$ for all layers need to be calculated only once.

\subsubsection{Adaptive Cross-view Sparse Alignment Learning}

This module fuses the aligned view-specific representations into a comprehensive representation. Traditional dense fusion forces low-dimensional views to generate meaningless ``pseudo-dimensions'' to match high-dimensional ones. To prevent this, we introduce a Multi-view Sparse Batch Normalization (MSBN) layer. By learning sparse cross-view correlations, MSBN adaptively suppresses redundant dimensions before fusion, ensuring each view contributes only its genuine information.

Before the fusion stage, each encoder consists of multiple fully connected layers, with Batch Normalization (BN) applied after each layer to mitigate covariate shift and enforce statistical alignment across heterogeneous views. For the $c$-th neuron in the $l$-th layer of the $v$-th encoder, the BN layer is defined as:
\begin{equation} \label{eq:5}
    z_{n,l+1,c}^v = \gamma_{l,c}^v \frac{z_{n,l,c}^v - \mu_l^v}{\sqrt{(\sigma_l^v)^2 + \epsilon}} + \beta_{l,c}^v,
\end{equation}
where $\gamma_{l,c}^v$ and $\beta_{l,c}^v$ are learnable scaling and shifting parameters,and $\epsilon$ represents a small scalar added to prevent numerical instability. The batch mean $\mu_l^v$ and variance ${\sigma_l^{v}}^2$ are computed as:
\begin{subequations} \label{eq:6}
\begin{align}
    \mu_l^v &= \frac{\sum_{n=1}^{N^b} \sum_{c=1}^{D_l^v} z_{n,l,c}^v}{N^b \cdot D_l^v}, \label{eq:6a} \\
    (\sigma_l^v)^2 &= \frac{\sum_{n=1}^{N^b} \sum_{c=1}^{D_l^v} (z_{n,l,c}^v - \mu_l^v)^2}{N^b \cdot D_l^v}. \label{eq:6b}
\end{align}
\end{subequations}
When the view-specific representations $\{\boldsymbol{z}_n^v\}$ are obtained, the next step is to fuse them into a comprehensive representation.
A common strategy in Multi-view Aligned Representation Learning (MARL)  assigns a unified weight $w^v$ to all dimensions within each view and leverages the fused comprehensive representation to guide the learning of individual views. However, in real-world scenarios, the dimensional discrepancies across views can be substantial, with high-dimensional views often containing information that low-dimensional views cannot capture. When the fused representation feeds back into the encoders, low-dimensional views are forced to generate "pseudo-dimensions" that they do not intrinsically possess in order to align with this high-dimensional information. Under the uniform weighting scheme, these pseudo-dimensions are treated as equivalent to genuine features. This forces the low-dimensional views to sacrifice the effective modeling of their original information and gradually lose their uniqueness, ultimately causing the multi-view learning to degenerate into a system dominated by the high-dimensional views.  

To address the issue of redundant dimensions, we introduce a Multi-view Sparse Batch Normalization (MSBN) layer on the aligned representations $\boldsymbol{z}_n^v$. We impose an $\ell_1$-norm penalty on the scaling factors $\{\gamma_c^v\}$ to enforce sparsity:
\begin{equation} \label{eq:7}
\mathcal{R}_{\text{sparse}} = \sum_{v=1}^V \sum_{c=1}^H |\gamma_c^v|, 
\end{equation}
where a value of $\gamma_c^v \to 0$ means the $c$-th dimension of $\boldsymbol{z}_n^v$ is effectively removed from the final representation. This mechanism allows the model to automatically identify and suppress redundant dimensions that are generated merely for alignment. The final comprehensive representation is computed by averaging the view-specific features:
\begin{equation} \label{eq:8}
\boldsymbol{z}_n = \frac{1}{V} \sum_{v=1}^V \boldsymbol{z}_n^v.  
\end{equation}
In the fused representation $\boldsymbol{z}_n$, a feature is activated if it appears in at least one view. Its final value simply depends on how many views share it. This means shared features will have higher values than features unique to a single view.

\subsection{Loss Function}
In this module, we design a self-supervised loss function to train the network. We first use the K-nearest neighbor (KNN) method to build a balanced similarity graph $\boldsymbol{S}^v$ for each view. We use this graph to construct pseudo-labels, which provide the supervisory signal for contrastive learning. Finally, we combine this contrastive loss with the sparsity penalty from the MSBN layer to end-to-end optimize the entire network.

\indent In particular, for each view, we construct a similarity graph $S^v$ where the nodes represent the data points. For any pair of nodes, an edge exists if they have the same class label or if one is among the $K$ closest neighbors of the other. The similarity score $s_{ij}^v$ between two nodes is defined as:
\begin{equation} \label{eq:10}
s_{ij}^v = \begin{cases} \exp\left(-\frac{\|\boldsymbol{x}_i^v - \boldsymbol{x}_j^v\|_2^2}{2\sigma^2}\right), & \parbox[t]{.4\linewidth}{$\text{if } \boldsymbol{x}_i^v \in \mathcal{N}_K(\boldsymbol{x}_j^v) \ \text{or}$ \\ $\boldsymbol{x}_j^v \in \mathcal{N}_K(\boldsymbol{x}_i^v)$,} \\ 0, & \text{otherwise}, \end{cases} 
\end{equation}
where $s_{ij}^v$ is the $(i, j)$-th entry representing the similarity between $\boldsymbol{x}_i^v$ and $\boldsymbol{x}_j^v$. $\mathcal{N}_K(\boldsymbol{x}_j^v)$ denotes the $K$-nearest neighbor set of $\boldsymbol{x}_j^v$. The scaling factor $\sigma = \frac{1}{n} \sum_i \| \boldsymbol{x}_i^v - \boldsymbol{x}_{i,k}^v \|_2^2$ is determined by the average distance to the $k$-th neighbor $\boldsymbol{x}_{i,k}^v$.

To capture the global structure across all views, we compute a unified consensus matrix $\boldsymbol{S}$. This is achieved by learning an optimal weighted combination of the individual matrices $\{\boldsymbol{S}^v\}$:
\begin{equation} \label{eq:10_opt}
\min_{\mathbf{S}, \boldsymbol{\alpha}^\top \mathbf{1} = 1, \boldsymbol{\alpha} \geq 0} \left\| \mathbf{S} - \sum_{v=1}^V \alpha^v \mathbf{S}^v \right\|_F^2,  
\end{equation}
where $\alpha^v$ represents the weight of the $v$-th view. The consensus matrix $\boldsymbol{S}$ reflects the unified neighborhood structure. A common approach is to employ a Graph Embedding (GE) loss based on $\{\boldsymbol{S}^v\}$ to guide the unsupervised learning process:
\begin{equation} \label{eq:11}
\mathcal{L}_{GE} = \frac{1}{2N} \sum_{i,j=1}^N \|\boldsymbol{z}_i - \boldsymbol{z}_j\|_2^2 S_{ij} + \lambda_2 \sum_{v,c} |\gamma_c^v|.
\end{equation}

This loss encourages representations of similar instances to be close.  However, this paradigm exhibits high sensitivity to the batch size (i.e., graph scale) during training. To improve stability, we adopt a contrastive learning loss instead. We generate binary pseudo-labels from the consensus matrix $\boldsymbol{S}$: if the similarity $s_{i,j}$ is greater than a threshold $\epsilon$, the pair is positive ($l_{ij} = 1$); otherwise, it is negative ($l_{ij} = 0$).
\begin{equation} \label{eq:12}
l_{ij} = \begin{cases} 1, & S_{ij} > \epsilon, \\ 0, & \text{otherwise}, \end{cases} 
\end{equation}

To optimize the view-specific encoders and sparse connections, we employ a self-supervised contrastive learning approach. This strategy encourages the model to minimize the distance between positive pairs while pushing negative pairs apart. The overall loss is expressed as:
\begin{equation} \label{eq:13}
\begin{split}
\min_{\{\boldsymbol{\Theta}_f^v\}} \frac{1}{2N^b} \sum_{i,j=1}^{N^b} \Big[ &(1 - l_{ij}) \max(m - \|\boldsymbol{z}_i - \boldsymbol{z}_j\|_2, 0)^2 \\
&+ l_{ij} \|\boldsymbol{z}_i - \boldsymbol{z}_j\|_2^2 \Big] + \lambda_2 \sum_{v,c} |\gamma_c^v|,
\end{split}
\end{equation}
where $\lambda_2$ denotes a hyper-parameter. The margin $m$ acts as a distance threshold. Specifically, the model only penalizes negative pairs whose Euclidean distance is less than $m$. Under this self-supervised framework, AdaMus learns unified multi-view representations from unlabeled data, which can be applied in downstream tasks such as classification and segmentation.

\begin{algorithm}[tb]
\caption{Pseudocode for AdaMuS.}
\label{alg:adamus}
\begin{algorithmic}[1]
\STATE \textbf{Input:} Multi-view dataset $\mathcal{X} = \{x_n^v\}$, pre-training epochs $E_{pre}$, fine-tuning epochs $E_{fine}$.
\STATE \textbf{Output:} Pruned network parameters $\{\Theta_f^{v'}\}_{v=1}^V$ and representations $\{z_n\}_{n=1}^N$.

\STATE ---------\textbf{Pre-training}---------
\STATE \textit{/Stage-1 Graph construction and encoder pre-training/}
\STATE Construct unified consensus matrix $S$ and pseudo-labels $\{l_{ij}\}$ by Eq.~14 and Eq.~16;
\FOR{$epoch = 1 : E_{pre}$}
    \STATE Compute aligned representations $z_n^v$ and MSBN scaling factors $\{\gamma_c^v\}$;
    \STATE Update initial encoders $\{\Theta_f^v\}_{v=1}^V$ via self-supervised contrastive loss by Eq.~17;
\ENDFOR

\STATE ---------\textbf{Adaptive Pruning}---------
\STATE \textit{/Stage-2 Parameter pruning via PNA/}
\FOR{each view $v$ and layer $l$}
    \STATE Calculate the specific pruning rate $\gamma_l^v$ via Wasserstein distance by Eq.~6;
    \STATE Adjust $\gamma_l^v$ with the view's unbalanced degree $\Lambda^v$ by Eq.~7 and Eq.~8;
    \STATE Prune the top $\gamma_l^v \times D_l^v$ redundant neurons based on their correlations;
\ENDFOR
\STATE Obtain the pruned encoders with parameters $\{\Theta_f^{v'}\}_{v=1}^V$;

\STATE ---------\textbf{Fine-tuning}---------
\STATE \textit{/* Stage-3 Sparse fusion \& fine-tuning/}
\FOR{$epoch = 1 : E_{fine}$}
    \STATE Extract pruned features $z_n^v$ using $\{\Theta_f^{v'}\}_{v=1}^V$ by Eq.~5;
    \STATE Suppress redundant dimensions via MSBN sparsity penalty by Eq.~11;
    \STATE Fuse into comprehensive representation $z_n$ by Eq.~12;
    \STATE Fine-tune parameters $\{\Theta_f^{v'}\}_{v=1}^V$ by gradient descent with the loss by Eq.~17;
\ENDFOR

\STATE \textbf{return} $\{\Theta_f^{v'}\}_{v=1}^V$ and $\{z_n\}_{n=1}^N$;
\end{algorithmic}
\end{algorithm}

\section{Experiments}

\subsection{Toy Example: }
\label{sec:toy_example}

\begin{figure*}[htbp]

    \centering

    \includegraphics[width=0.95\textwidth]{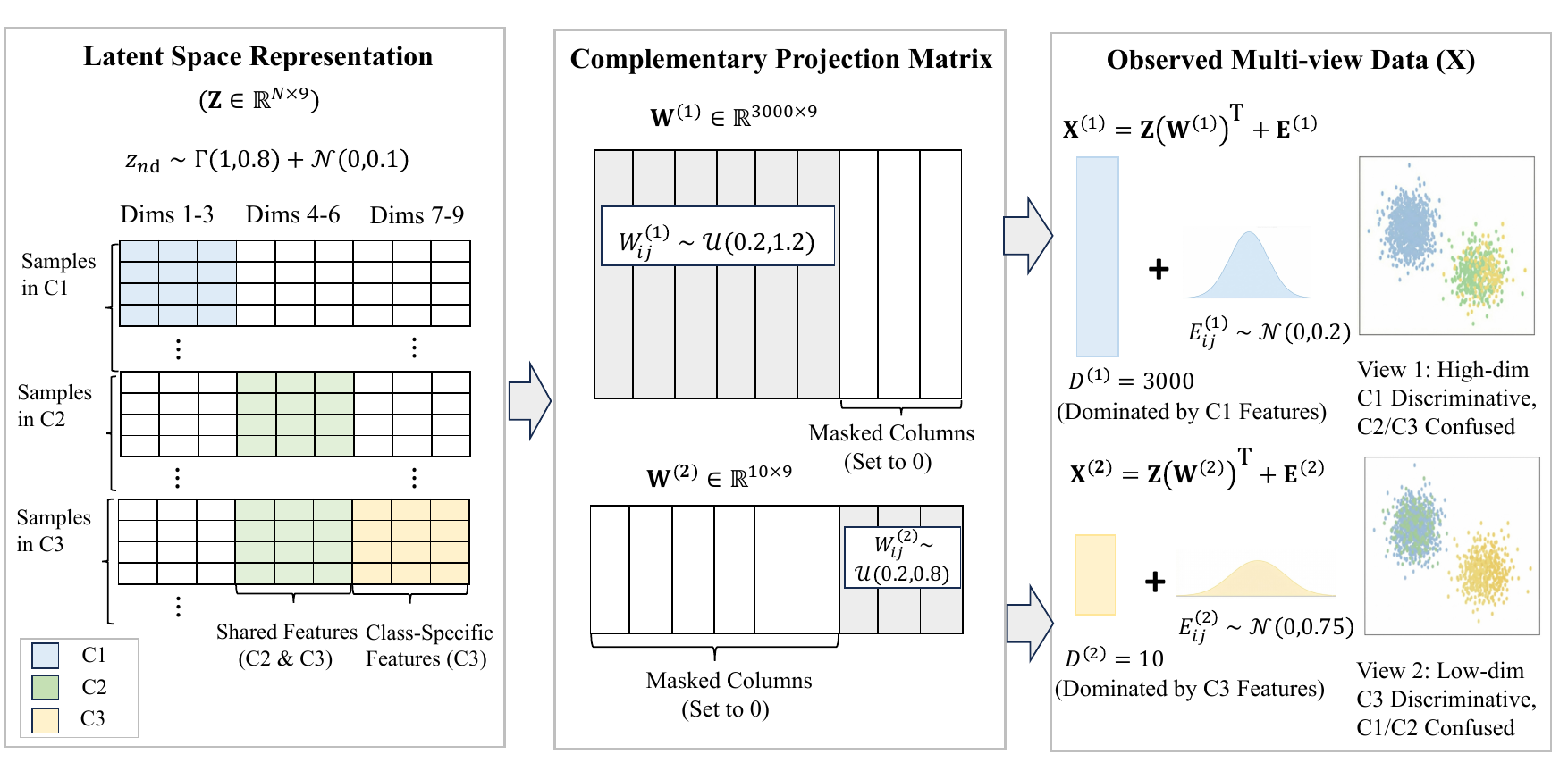}

    \caption{Construction process of the multi-view toy example.}

    \label{fig:toy_example}

\end{figure*}

We assess the performance of AdaMuS through clustering, classification, and segmentation tasks. The experiments use one toy example and seven real-world datasets, covering both balanced and unbalanced scenarios.

   
    
    

To investigate multi-view learning under extreme dimensional disparity, we categorize existing fusion strategies into three paradigms: (1)\textbf{Direct Concatenation}, (2)\textbf{High-to-Low Compression}, and (3) \textbf{Shared Projection}. While effective in balanced scenarios, they struggle under extreme disparity. Thus, we construct a Toy Example to empirically demonstrate their limitations and our approach's effectiveness, guided by two Research Questions (\textbf{RQs}):

\begin{itemize}
    \item \textbf{RQ1: What are the limitations of existing paradigms in unbalanced multi-view learning?} 
    \begin{enumerate}
        \item[(1)] \textbf{Dominance Issue:} In direct concatenation, does the high-dimensional view overwhelm the low-dimensional view?
        \item[(2)] \textbf{Information Loss:} In high-to-low compression, does the drastic dimensionality reduction result in the loss of critical information?
        \item[(3)] \textbf{Overfitting Risk:} When projecting the low-dimensional view to a higher space, does it suffer from overfitting due to redundant parameters?
    \end{enumerate}

    \item \textbf{RQ2: Can AdaMus effectively handle the dimensional imbalance?} 
    \begin{enumerate}
        \item[(1)] \textbf{Discriminative Capability:} 
        Can AdaMus successfully capture the essential features of each class?
        
        \item[(2)] \textbf{Overfitting Prevention:} 
        Can AdaMus avoid the overfitting problem when projecting the low-dimensional view into a higher-dimensional space?
        
        \item[(3)] \textbf{Sparse Alignment:} 
        Can AdaMus capture the sparse  correlations?
    \end{enumerate}
\end{itemize}

\begin{figure*}[t]
    \centering
    \setlength{\fboxsep}{1pt} 
    \setlength{\fboxrule}{0.5pt}
    
    \subfloat[View 1: High-Dim \protect\\ (Acc: 65.73\% $\pm$ 2.40\%)]{%
        \fbox{\includegraphics[width=0.28\textwidth]{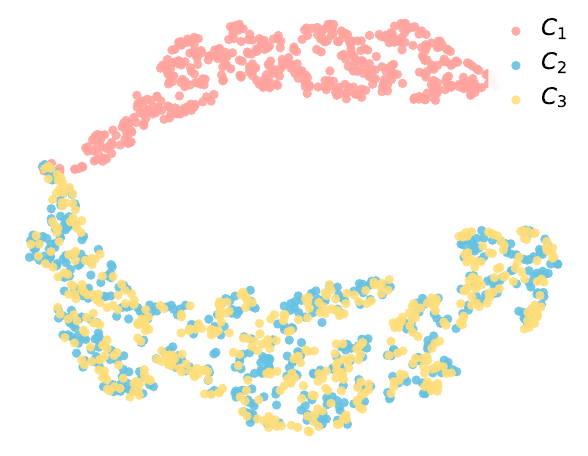}}%
    }\hfill
    \subfloat[View 2: Low-Dim \protect\\ (Acc: 63.33\% $\pm$ 2.28\%)]{%
        \fbox{\includegraphics[width=0.28\textwidth]{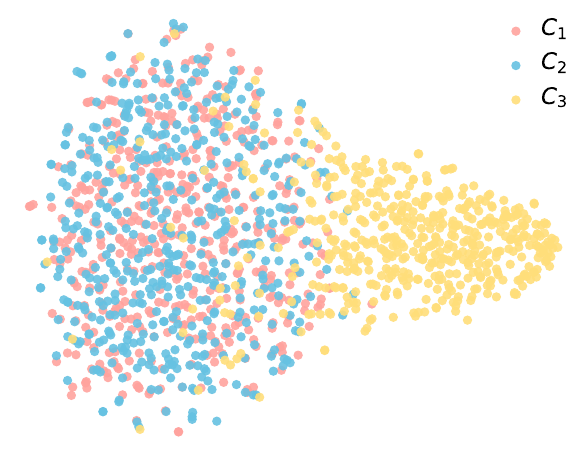}}%
    }\hfill
    \subfloat[Naive Concat \protect\\ (Acc: 67.20\% $\pm$ 2.35\%)]{%
        \fbox{\includegraphics[width=0.28\textwidth]{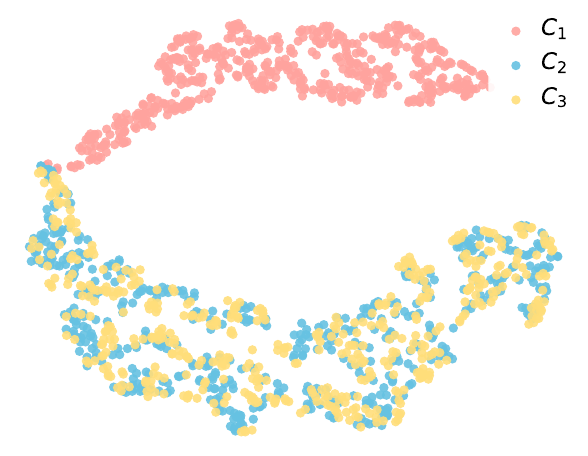}}%
    }
    
    \vspace{0.2cm} 
    
    \subfloat[High-to-low compression \protect\\ (Acc: 67.67\% $\pm$ 2.05\%)]{%
        \fbox{\includegraphics[width=0.28\textwidth]{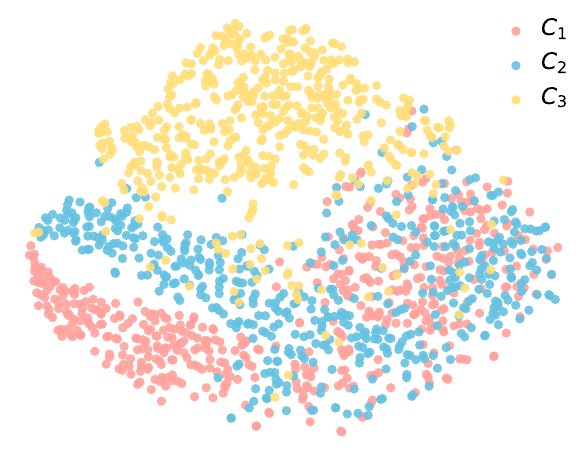}}%
    }\hfill
    \subfloat[Inter. Dim Map \protect\\ (Acc: 87.20\% $\pm$ 2.04\%)]{%
        \fbox{\includegraphics[width=0.28\textwidth]{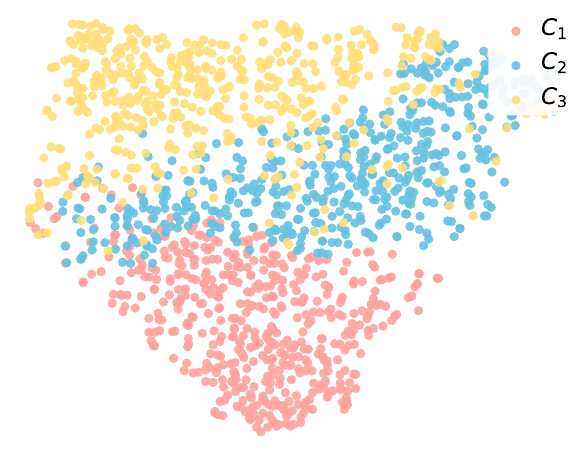}}%
    }\hfill
    \subfloat[AdaMus (Ours) \protect\\ (Acc: \textbf{97.27\%} $\pm$ 0.33\%)]{%
        \fbox{\includegraphics[width=0.28\textwidth]{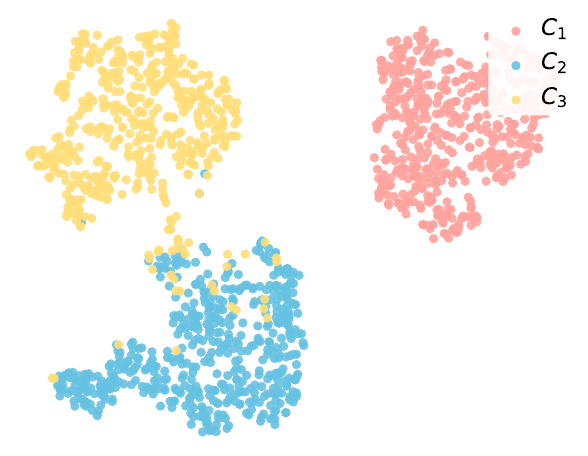}}%
    }
    
    \caption{t-SNE visualization of the learned representations on the toy example.}
    \label{fig:tsne}
\end{figure*}

\vspace{0.2cm}
\noindent \textbf{Data Construction Details.} 
Specifically, the constructed dataset consists of $N=1500$ samples evenly distributed across 3 categories ($C_1, C_2, C_3$). 
The underlying latent space $\mathbf{Z} \in \mathbb{R}^{N \times 9}$ represents the ground truth semantic information. 
Each latent code is sampled from a Gamma distribution $\Gamma(1, 0.8)$ superimposed with Gaussian noise $\mathcal{N}(0, 0.1)$. 
The semantic structure is defined as follows: dimensions 1-3 are active for $C_1$, dimensions 4-6 for $C_2$, and dimensions 7-9 for $C_3$. 
Consequently, $C_2$ and $C_3$ share features in dimensions 4-6, making them distinguishable only via dimensions 7-9.

To simulate extreme dimensional disparity, we generate two views via linear projection $\mathbf{X}^{(v)} = \mathbf{Z}(\mathbf{W}^{(v)})^T + \mathbf{E}^{(v)}$. 
{View 1} is high-dimensional ($\mathbf{X}^{(1)} \in \mathbb{R}^{N \times 3000}$) with a projection matrix $\mathbf{W}^{(1)}$ initialized from a uniform distribution $\mathcal{U}(0.2, 1.2)$. 
{View 2} is extremely low-dimensional ($\mathbf{X}^{(2)} \in \mathbb{R}^{N \times 10}$) with $\mathbf{W}^{(2)} \sim \mathcal{U}(0.2, 0.8)$. 
Crucially, we implement structural masking to enforce complementarity: 
we set the last 3 columns of $\mathbf{W}^{(1)}$ to zero, rendering View 1 blind to the unique features of $C_3$, thereby causing confusion between $C_2$ and $C_3$. 
Next, we set the first 6 columns of $\mathbf{W}^{(2)}$ to zero, meaning View 2 can only recognize class $C_3$. 
Finally, independent Gaussian noise is injected into both views, parameterized as $\mathbf{E}^{(1)} \sim \mathcal{N}(0, 0.2)$ and $\mathbf{E}^{(2)} \sim \mathcal{N}(0, 0.75)$.

\vspace{0.2cm}
\noindent \textbf{T-SNE Visualization Analysis.} 
The visualization results in Fig.~\ref{fig:tsne} provide empirical evidence for our assumptions. 
First, as shown in Fig.~\ref{fig:tsne}(a) and (b), each single view successfully identifies specific categories while confusing others, confirming that both views contain unique and complementary information.
However, simple fusion strategies fail to preserve this complementarity: 
the naive concatenation (Fig.~\ref{fig:tsne}c) fails to capture the unique patterns of the low-dimensional view, indicating that the informative low-dimensional signals are overwhelmed by the high-dimension. 
Similarly, the high-to-low compression (Fig.~\ref{fig:tsne}d) exhibits blurred class boundaries, demonstrating that forcibly reducing dimensions leads to a significant loss of effective information. 
In contrast, our AdaMus method (Fig.~\ref{fig:tsne}f) generates the most distinct clusters, proving its ability to effectively integrate these complementary semantics despite the dimensional disparity.

\begin{figure}[t]
    \centering
    \includegraphics[width=0.7\linewidth]{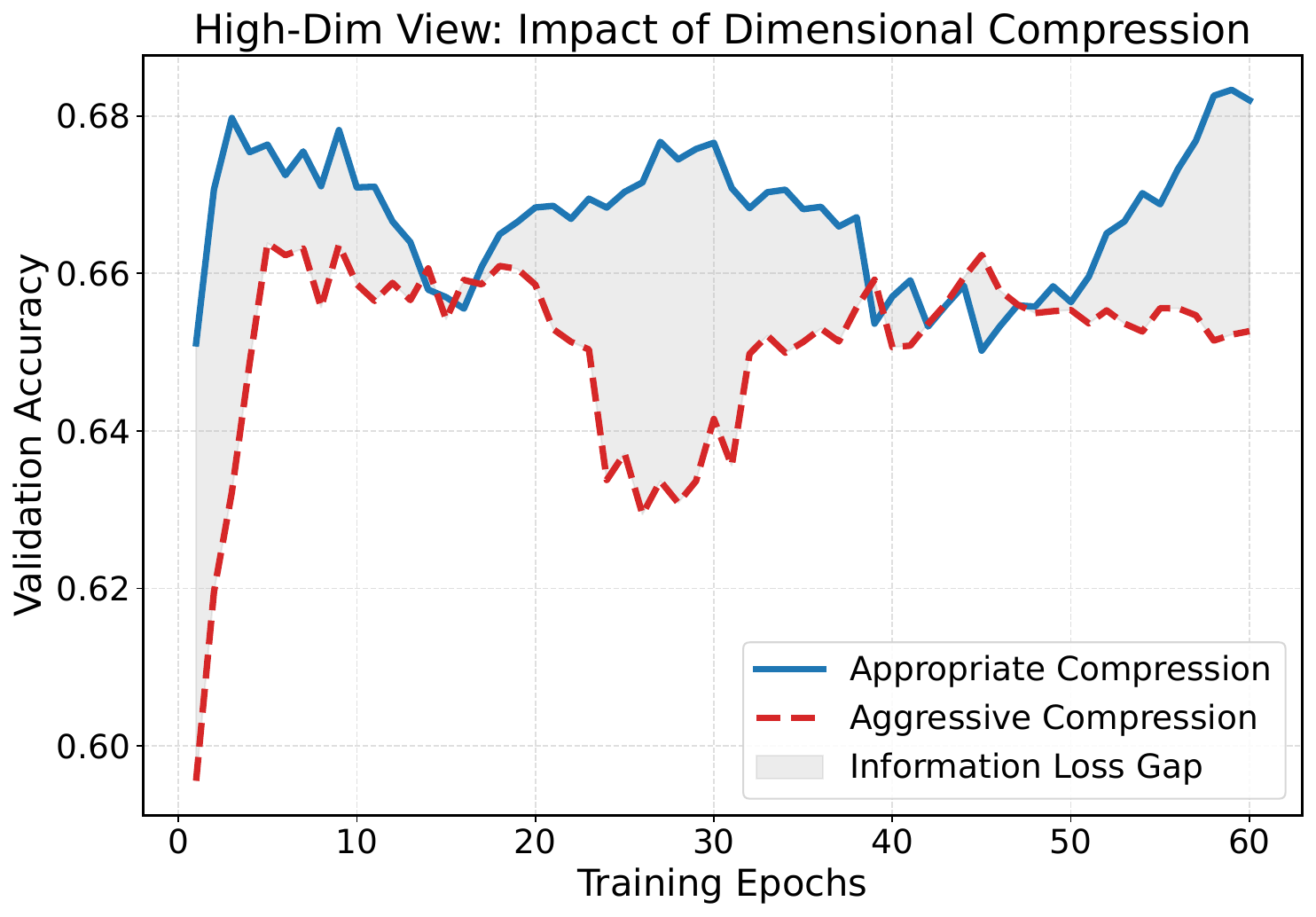}
    \vspace{-8pt} 
    \caption{\textbf{Impact of Dimensional Compression.} Comparison between appropriate compression ($3000 \to 1024$) and over-compression ($3000 \to 128$). The gray gap highlights the significant information loss caused by aggressive compression, verifying the need for a high-dimensional subspace.}
    \label{fig:dim_compression}
\end{figure}

\noindent\textbf{Impact of Dimensional Compression.} 
We investigate whether compressing the high-dimensional view into a constrained space induces information loss. 
We compare two projection strategies for View 1: 
an appropriate compression architecture ($3000 \to 1024 \to 512$) and an aggressive compression architecture ($3000 \to 128 \to 64$). 
Figure~\ref{fig:dim_compression} demonstrates that the appropriate compression strategy significantly outperforms the aggressive one. 
The result indicates that excessive dimensionality reduction leads to the loss of high-dimensional information.

\begin{figure}[t]
    \centering
    \begin{minipage}[b]{0.495\linewidth}
        \centering
        \includegraphics[width=\linewidth]{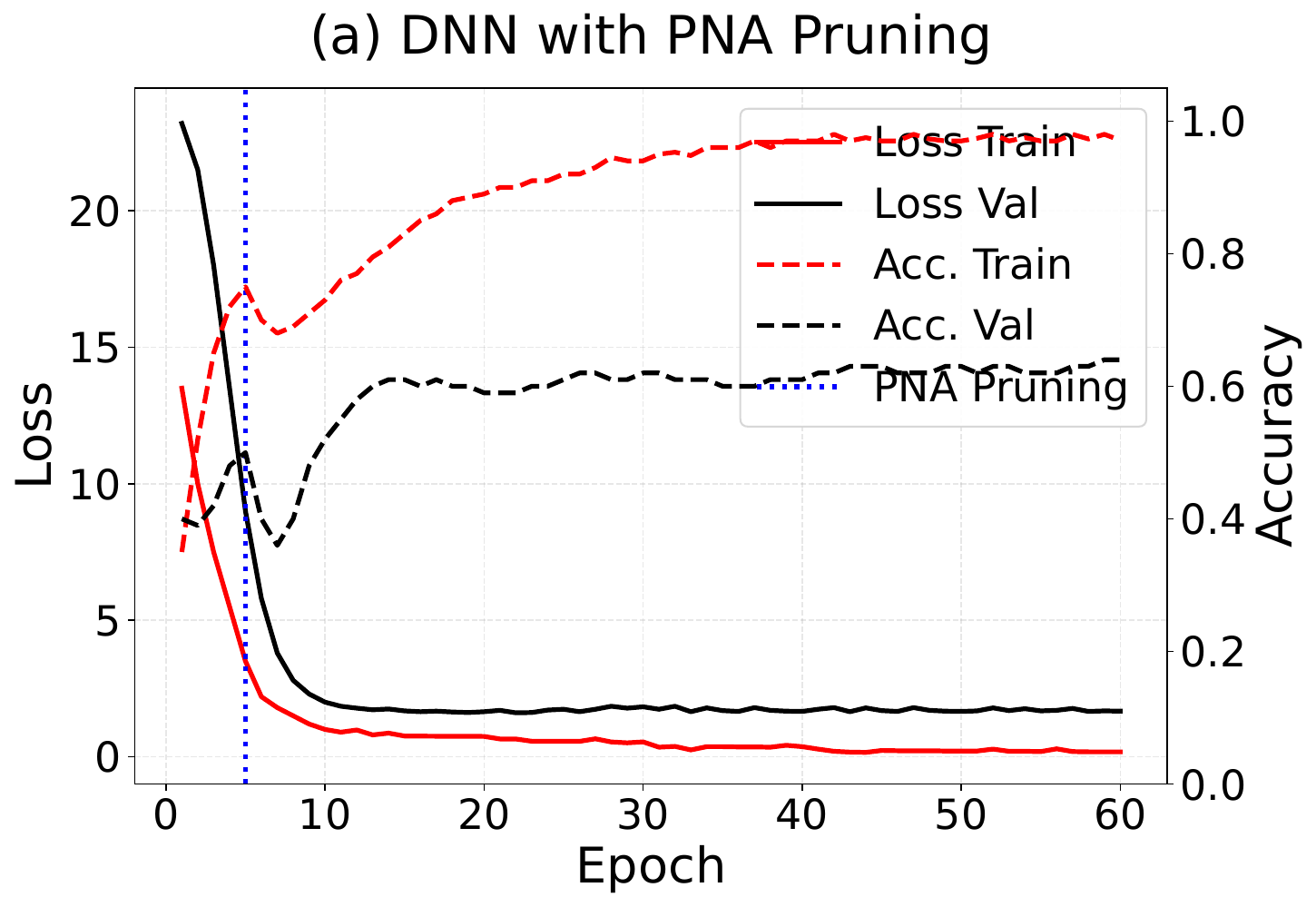}
        \vspace{-5pt} 
        \centerline{\small (a) PNA Pruning}
    \end{minipage}
    \hfill 
    \begin{minipage}[b]{0.495\linewidth}
        \centering
        \includegraphics[width=\linewidth]{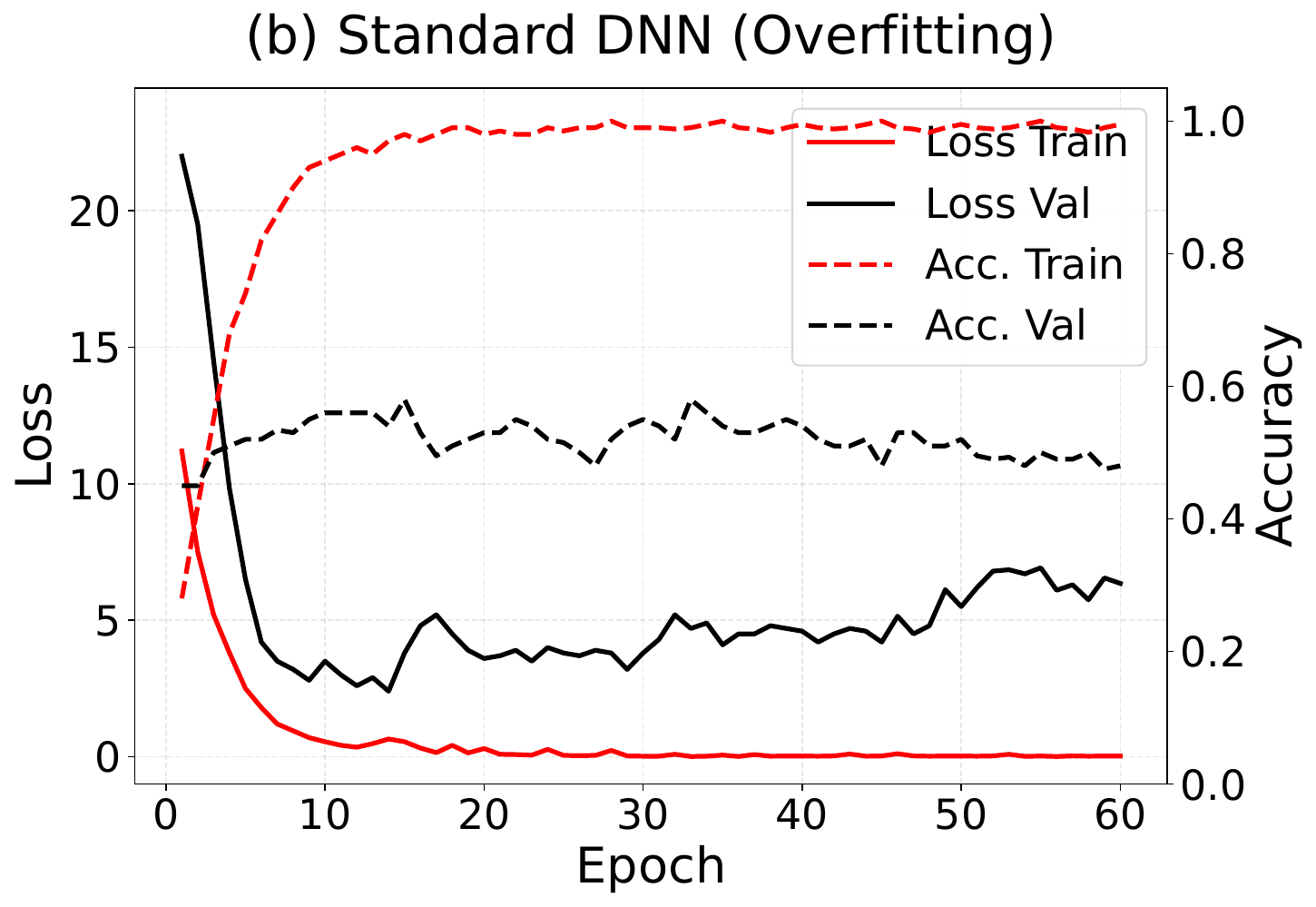}
        \vspace{-5pt} 
        \centerline{\small (b) Standard DNN}
    \end{minipage}
    
    \vspace{5pt} 
    \caption{\textbf{Validation of the PNA pruning module.} We compare the learning dynamics of (a) PNA-enhanced DNN vs. (b) Standard DNN during dimensional expansion ($10 \to 512 \to 1024$). The Standard DNN suffers from overfitting, whereas PNA ensures stable convergence by pruning redundant connections.}
    \label{fig:overfitting}
\end{figure}

\noindent\textbf{Overfitting in Expansion and PNA Mitigation.} 
We investigate overfitting during low-dimensional expansion and the PNA module's effectiveness. We compare a Standard DNN ($10 \to 1024 \to 512$) against a PNA-enhanced DNN, which applies pruning after 5 epochs of pre-training followed by fine-tuning. As shown in Fig.~\ref{fig:overfitting}(b), the Standard DNN exhibits typical overfitting: while the training accuracy is high, the validation accuracy fluctuates violently, and the validation loss shows a distinct upward trend. In contrast, Fig.~\ref{fig:overfitting}(a) demonstrates that PNA achieves stable accuracy and consistently decreasing loss. This confirms that PNA effectively mitigates the overfitting caused by expanding the low-dimensional view.

\begin{figure}[t] 
    \centering
    \includegraphics[width=\linewidth]{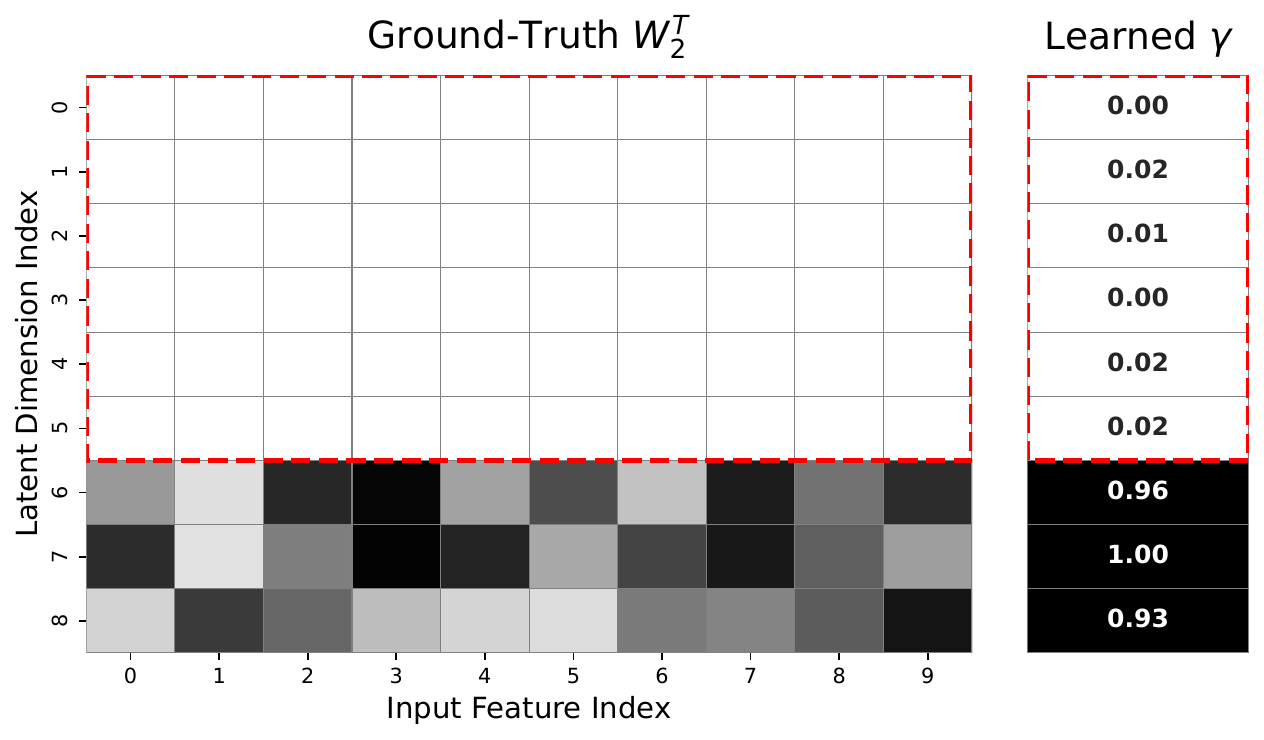}
    \caption{Verification of sparse alignment. (Left) Ground-truth mapping matrix $W_2^T$, where red dashed boxes indicate irrelevant dimensions. (Right) The learned scaling factor $\boldsymbol{\gamma}$ effectively suppresses these dimensions while activating the informative ones.}
    \label{fig:sparse_complementarity}
\end{figure}

\noindent\textbf{Verification of Sparse Alignment.} 
To verify the sparse alignment capability, we compare the ground-truth matrix regarding View 2 with the learned scaling factor $\boldsymbol{\gamma}$ in Fig.~\ref{fig:sparse_complementarity}. 
The red dashed box in the left panel highlights the dimensions that are irrelevant to this view. 
As observed in the right panel, the learned weights effectively suppress these irrelevant dimensions while activating the informative ones. 
This high consistency confirms that AdaMus successfully identifies the correct feature correspondence without supervision.

\begin{table}[t]
\centering
\caption{DATASET STATISTICS. $\Lambda$ INDICATES THE UNBALANCED DEGREE. NYUV2 IS EMPLOYED FOR SEGMENTATION.}
\label{tab:dataset_summary}

\resizebox{\columnwidth}{!}{%
\begin{tabular}{l | c | c | c | c}
\toprule
Dataset & Size & Category & Dimensions & $\Lambda$ \\
\midrule
UCI       & 1600 & 10           & $6/47/240$                                 & 2.88 \\
CUB       & 480  & 10           & $30/1024$                                  & 3.22 \\
ORL       & 400  & 40           & $30/3304/6750$                             & 2.54 \\
MSRCV1    & 210  & 7            & $20/100/210/256/512/1302$                  & 2.41 \\
Mfeat     & 2000 & 10           & $240/216/76/64/47/6$                       & 2.22 \\
100leaves & 1600 & 100          & $5/64/64$                                  & 2.14 \\
DEAP      & 1280 & 2            & $1.5 \times 10^5 / 32 / 8$                 & 4.07 \\
NYUv2     & 1449 & $\backslash$ & $3 \times 640 \times 480 / 640 \times 480$ & 1.71 \\
\bottomrule
\end{tabular}%
}
\end{table}
\begin{table*}[t] 
    \centering 
    \caption{Clustering performance comparison on different datasets (mean result $\pm$ standard deviation). AdaMuS is our proposed method. The best and the second best results are highlighted by \textbf{boldface} and \underline{underlined}, respectively.} 
    \label{pcoc_journal_style_final} 
    \setlength{\tabcolsep}{2.5pt} 
    \resizebox{\textwidth}{!}{ 
    \begin{tabular}{ll|ccccccccccccc} 
        \toprule 
        \multirow{2}{*}{\textbf{Datasets}} & \multirow{2}{*}{\textbf{Metric}} & \multicolumn{13}{c}{\textbf{Methods}} \\ 
        \cmidrule(lr){3-15} 
        & & Concat & DCCAE & CGCN & BMGC & ICMVC & CANDY & GUMRL & PDMF & UIMC & SURER & STCMC-UR & UMDL & \textbf{AdaMuS} \\ 
        \midrule 
        \multirow{3}{*}{UCI} & ACC & 71.56±3.98 & 74.62±1.19 & 70.00±1.96 & 73.45±1.68 & 85.70±1.18 & 69.18±2.05 & 86.09±0.08 & 85.20±1.09 & 76.88±2.01 & 77.90±1.12 & 72.24±1.39 & \underline{87.51±0.32} & \bf{91.49±1.08} \\ 
        & NMI & 72.39±2.11 & 79.77±1.11 & 78.01±0.87 & 73.33±1.57 & \underline{83.13±1.05} & 67.23±2.14 & 78.66±0.91 & 75.84±1.44 & 67.15±2.33 & 75.57±1.76 & 73.27±0.60 & 79.08±1.02 & \bf{86.93±0.97} \\ 
        & ARI & 62.28±3.15 & 71.97±0.78 & 57.71±2.96 & 63.64±2.83 & \underline{75.14±1.31} & 57.61±1.75 & 69.63±0.26 & 71.51±1.06 & 60.44±3.11 & 67.06±2.17 & 62.46±1.10 & 74.23±0.14 & \bf{82.85±0.90} \\ 
        \midrule 
        \multirow{3}{*}{CUB} & ACC & 64.23±2.91 & 69.71±1.83 & 57.08±2.42 & 68.96±1.95 & 62.53±2.40 & 51.48±2.90 & 61.73±2.16 & 70.47±3.71 & 70.11±1.17 & 69.60±1.02 & 55.22±2.55 & \underline{80.12±1.64} & \bf{81.65±1.74} \\ 
        & NMI & 60.85±1.69 & 67.34±2.54 & 69.29±1.34 & 57.46±2.85 & 54.07±1.76 & 64.75±2.31 & 53.05±1.93 & 66.83±1.71 & 65.82±1.07 & 72.29±0.99 & 63.55±2.63 & \underline{73.43±2.74} & \bf{77.21±0.45} \\ 
        & ARI & 57.45±2.05 & 64.01±1.47 & 53.31±2.01 & 58.20±2.55 & 44.46±1.92 & 43.89±1.55 & 43.94±0.74 & 59.29±1.91 & \bf{75.44±1.58} & 64.85±0.95 & 45.39±3.12 & 68.84±0.50 & \underline{73.75±0.65} \\ 
        \midrule 
        \multirow{3}{*}{ORL} & ACC & 61.50±1.81 & 68.07±1.31 & 59.57±2.05 & 49.60±2.95 & 72.25±1.73 & \underline{76.50±1.85} & 76.15±1.06 & 64.70±2.76 & 66.17±2.93 & 70.50±1.55 & 52.00±1.34 & 69.65±1.53 & \bf{80.98±1.55} \\ 
        & NMI & 76.22±0.84 & 82.47±1.25 & 67.52±0.67 & 63.52±2.12 & 83.42±1.21 & 86.65±1.02 & \bf{89.72±0.22} & 71.62±2.08 & 80.15±1.00 & 84.25±0.85 & 64.41±1.16 & 83.64±1.04 & \underline{87.37±0.82} \\ 
        & ARI & 64.96±1.23 & 62.14±1.32 & 53.24±2.64 & 41.69±1.68 & 61.69±1.55 & 65.70±1.04 & 70.05±0.10 & 56.25±2.85 & 51.30±2.57 & \underline{72.38±1.20} & 43.32±1.61 & 65.95±0.69 & \bf{74.95±0.40} \\ 
        \midrule 
        \multirow{3}{*}{MSRCV1} & ACC & 69.38±3.93 & 78.24±1.45 & 81.87±1.31 & 55.57±2.53 & 66.19±2.15 & 85.52±1.52 & 88.05±2.68 & 90.47±1.71 & 80.76±3.16 & \underline{93.81±0.24} & 91.69±1.31 & 94.68±0.38 & \bf{95.05±0.77} \\ 
        & NMI & 59.34±3.35 & 71.67±1.85 & 74.24±1.48 & 51.28±1.83 & 64.56±1.91 & 81.50±0.02 & 88.67±0.60 & 78.86±1.43 & 70.64±4.42 & \underline{91.86±1.43} & 87.74±1.45 & 88.93±1.13 & \bf{93.07±0.76} \\ 
        & ARI & 62.15±1.07 & 62.78±1.52 & 67.29±2.05 & 53.23±2.33 & 52.91±2.04 & 75.10±0.03 & 79.00±0.14 & 73.86±1.03 & 64.13±4.31 & \underline{89.40±0.45} & 84.69±1.86 & 84.98±0.39 & \bf{91.75±0.60} \\ 
        \midrule 
        \multirow{3}{*}{Mfeat} & ACC & 60.95±2.81 & 77.25±1.62 & 93.50±0.73 & 63.55±2.31 & 87.74±1.02 & 72.30±0.63 & 91.27±0.03 & \underline{93.85±0.05} & 84.27±3.62 & 93.80±0.35 & 81.56±1.92 & 93.58±0.05 & \bf{95.03±0.05} \\ 
        & NMI & 57.69±2.79 & 79.67±1.48 & 90.67±1.08 & 55.93±2.94 & 86.91±0.58 & 69.12±2.35 & 92.21±0.08 & 93.62±0.07 & 83.24±1.26 & 88.32±0.28 & 84.11±2.26 & \underline{95.60±0.61} & \bf{96.06±0.08} \\ 
        & ARI & 50.59±3.15 & 68.34±3.72 & 89.50±0.92 & 42.48±2.87 & 81.05±1.15 & 57.61±2.10 & \underline{91.89±0.07} & 90.21±0.07 & 79.18±3.47 & 86.00±0.22 & 75.14±2.89 & 89.12±0.67 & \bf{93.90±0.05} \\ 
        \midrule 
        \multirow{3}{*}{100Leaves} & ACC & 64.44±2.17 & 73.70±5.86 & 59.62±3.37 & 65.00±1.37 & 60.70±3.77 & 48.76±3.09 & 88.30±1.42 & 62.76±4.77 & 57.70±0.44 & \underline{90.88±1.15} & 59.25±0.05 & 75.50±1.23 & \bf{93.61±1.65} \\ 
        & NMI & 77.11±0.89 & 77.67±2.26 & 76.30±2.12 & 64.02±1.94 & 74.22±1.34 & 62.37±2.15 & 96.13±0.35 & 83.83±2.16 & 75.84±0.48 & \underline{96.97±0.26} & 67.01±0.39 & 89.17±0.36 & \bf{97.81±0.36} \\ 
        & ARI & 52.87±1.44 & 65.28±5.06 & 42.16±3.06 & 51.35±2.31 & 48.37±1.96 & 37.56±2.84 & 85.37±1.62 & 47.02±6.55 & 42.26±1.00 & \underline{88.34±1.62} & 50.11±0.72 & 67.25±0.65 & \bf{92.99±1.12} \\ 
        \bottomrule 
    \end{tabular} 
    } 
\end{table*}

\begin{table*}[t]
\centering
\caption{Classification comparison across benchmarks (Mean $\pm$ Std). \textbf{Bold} and \underline{underlined} values denote the highest and second-highest scores, respectively.}
\label{classify_journal_style_updated}
\scalebox{0.85}{
\begin{tabular}{ll|ccccccc}
\toprule
\multirow{2}{*}{\textbf{Datasets}} & \multirow{2}{*}{\textbf{Metric}} & \multicolumn{6}{c}{\textbf{Methods}} \\
\cmidrule(lr){3-8}
& & PDMF & GUMRL & SURER & RCML & UMDL & \textbf{AdaMuS (Ours)} \\
\midrule
\multirow{2}{*}{UCI}
& ACC & 94.43$\pm$0.15 & 96.88$\pm$0.09 & 96.00$\pm$0.25 & 96.67$\pm$0.72 & \underline{97.25$\pm$0.17} & \textbf{98.50$\pm$0.03} \\
& F1  & 96.03$\pm$0.19 & 96.78$\pm$0.10 & 95.87$\pm$0.32 & 96.63$\pm$0.74 & \underline{97.00$\pm$0.17} & \textbf{98.50$\pm$0.03} \\
\midrule
\multirow{2}{*}{CUB}
& ACC & 87.29$\pm$0.69 & 85.42$\pm$1.16 & \underline{89.17$\pm$0.24} & 88.75$\pm$2.64 & 87.99$\pm$0.40 & \textbf{90.00$\pm$0.31} \\
& F1  & 87.17$\pm$0.58 & 85.50$\pm$0.84 & \underline{89.09$\pm$0.33} & 88.20$\pm$2.26 & 87.56$\pm$0.50 & \textbf{89.57$\pm$0.17} \\
\midrule
\multirow{2}{*}{ORL}
& ACC & 69.50$\pm$4.03 & 92.12$\pm$0.27 & 94.00$\pm$0.21 & \underline{96.62$\pm$0.72} & 92.25$\pm$0.43 & \textbf{97.25$\pm$0.02} \\
& F1  & 67.58$\pm$2.97 & 91.79$\pm$0.15 & 94.26$\pm$0.18 & \underline{95.84$\pm$1.19} & 92.03$\pm$0.59 & \textbf{97.20$\pm$0.01} \\
\midrule
\multirow{2}{*}{MSRCV1}
& ACC & 88.00$\pm$2.10 & 92.86$\pm$0.14 & 97.62$\pm$0.08 & 92.62$\pm$1.44 & \underline{98.00$\pm$0.31} & \textbf{99.52$\pm$0.01} \\
& F1  & 86.51$\pm$1.77 & 92.55$\pm$0.12 & 97.68$\pm$0.12 & 90.60$\pm$1.17 & \underline{98.04$\pm$0.35} & \textbf{99.52$\pm$0.02} \\
\midrule
\multirow{2}{*}{Mfeat}
& ACC & \underline{98.05$\pm$0.22} & 97.50$\pm$0.10 & 95.50$\pm$0.15 & 96.65$\pm$1.14 & 97.00$\pm$0.12 & \textbf{98.74$\pm$0.03} \\
& F1  & \underline{97.89$\pm$0.38} & 97.50$\pm$0.12 & 95.50$\pm$0.10 & 96.59$\pm$1.16 & 97.25$\pm$0.24 & \textbf{98.99$\pm$0.02} \\
\midrule
\multirow{2}{*}{100leaves}
& ACC & 83.34$\pm$2.48 & 90.34$\pm$0.39 & \underline{98.38$\pm$0.03} & 80.91$\pm$1.95 & 97.69$\pm$0.08 & \textbf{98.88$\pm$0.09} \\
& F1  & 79.68$\pm$2.01 & 89.75$\pm$0.67 & \underline{98.31$\pm$0.02} & 77.72$\pm$2.55 & 97.66$\pm$0.14 & \textbf{98.60$\pm$0.04} \\
\midrule
\multirow{2}{*}{DEAP}
& ACC & 92.95$\pm$1.31 & 88.25$\pm$1.87 & \underline{96.12$\pm$0.23} & 84.03$\pm$0.38 & 94.71$\pm$1.37 & \textbf{98.46$\pm$0.01} \\
& F1  & 92.74$\pm$1.35 & 87.94$\pm$1.90 & \underline{96.02$\pm$0.28} & 83.66$\pm$0.25 & 94.57$\pm$1.41 & \textbf{98.42$\pm$0.01} \\
\bottomrule
\end{tabular}
}
\end{table*}

\subsection{Experiments on Real-World Datasets}

\subsubsection{Datasets}
\label{sec:datasets}

We first outline the datasets utilized for the primary tasks. Summary statistics are provided in Table~\ref{tab:dataset_summary}.

For \emph{Clustering and Classification}, we employ six benchmark datasets. 
\textbf{UCI Multiple Features (UCI)}\footnote{https://archive.ics.uci.edu/ml/datasets/Multiple+Features} is a standard benchmark for digit recognition (0-9). It represents samples using three descriptors: 6D morphological features, 47D Zernike moments, and 240D pixel averages.
\textbf{Caltech-USD Birds (CUB)} contains 480 images spanning 10 fine-grained bird species. The original data includes a 1024-dimensional visual view and a 300-dimensional textual view. To simulate severe dimensional disparity, we apply Principal Component Analysis (PCA) to compress the textual view to 30 dimensions.
\textbf{ORL} consists of 400 face images from 40 subjects. It comprises three feature sets: 4096D intensity, 3304D Local Binary Patterns (LBP), and 6750D Gabor features. We specifically reduce the Gabor view to 30 dimensions to test performance under extreme imbalance.
\textbf{MSRCV1} includes 210 images across 7 object categories. It provides six distinct views (GIST, HOG, LBP, SIFT, CENT, and CMT). We manually reduce one 48-dimensional view to 20 dimensions to introduce structural imbalance.
Additionally, we evaluate on two complex datasets: \textbf{100Leaves} and \textbf{Multiple Features (Mfeat)}. 
The \textbf{100Leaves} dataset comprises 1,600 samples from 100 plant species, initially defined by three 64-dimensional views (shape, texture, margin). We induce imbalance by compressing the first view to 5 dimensions, yielding a configuration of [5, 64, 64].
The \textbf{Mfeat} dataset provides 2,000 samples for digits 0-9. It features six diverse views: 76D Fourier coefficients, 216D profile correlations, 64D Karhunen-Loeve coefficients, 240D pixel averages, 47D Zernike moments, and 6D morphological features. \textbf{DEAP} comprises 1,280 samples recording 32 participants' emotional responses. Unlike previous datasets, it presents a \emph{natural} extreme imbalance: the raw Video view operates in a high-dimensional pixel space ($\approx 1.5 \times 10^5$D), while the EEG and Peripheral views contain only 32D and 8D sensor readings, respectively.

For \emph{Segmentation}, we utilize the \textbf{NYU-Depth V2 (NYUv2)}\textsuperscript{4} dataset. This dataset captures various indoor scenes using RGB and Depth sensors from a Microsoft Kinect. The volumetric data is annotated with 13 semantic classes, such as `bed', `books', and `window'.

\subsubsection{Compared Methods}
\label{sec:compared_methods}
We evaluate AdaMuS against a comprehensive set of 13 baseline methods. We organize these competitors into five distinct categories according to their core integration strategies:

(1) \textbf{Decision-level fusion}: \textbf{ECML}~[16] resolves conflicts in multi-view data via an opinion aggregation mechanism. \textbf{STCMC-UR}~\cite{hu2025self} utilizes evidence theory to synthesize high-confidence pseudo-labels for feature refinement. \textbf{UIMC}~\cite{fang2021unbalanced} achieves a unified partition by aggregating the clustering outcomes derived from each view in isolation.

(2) \textbf{Direct concatenation}: \textbf{Concat} merges raw features from all views into a single vector for input.

(3) \textbf{Joint representation learning}: \textbf{DCCAE}~\cite{wang2015deep} learns a canonical representation shared across the views via deep autoencoder optimization. \textbf{PDMF}~\cite{xu2023aaai} leverages a pretraining stage to refine inter-view alignment.

(4) \textbf{Graph-based learning}: \textbf{SURER}~\cite{guo2024robust} captures sample topology through a graph learning module. \textbf{GUMRL}~\cite{zheng2022graph} exploits low-level geometric information to enhance clustering. \textbf{CGCN}~\cite{wang2024partially} focuses on aligning graph structures for scenarios with partial views.

(5) \textbf{Methods for view imbalance and imperfect data}: \textbf{BMGC}~\cite{shen2024balanced} handles structural imbalance by identifying and leveraging dominant views. \textbf{ICMVC}~\cite{chao2024incomplete} manages missing data using a high-confidence guidance strategy. \textbf{CANDY}~\cite{guo2024robust} mitigates the impact of noisy views via context mining and spectral decomposition. \textbf{UMDL}~\cite{xu2023unbalanced}, a state-of-the-art approach for dimensional imbalance, aligns views by constructing over-complete dictionaries for low-dimensional inputs.

\begin{figure}[!t]
\centering
\includegraphics[width=2.4in]{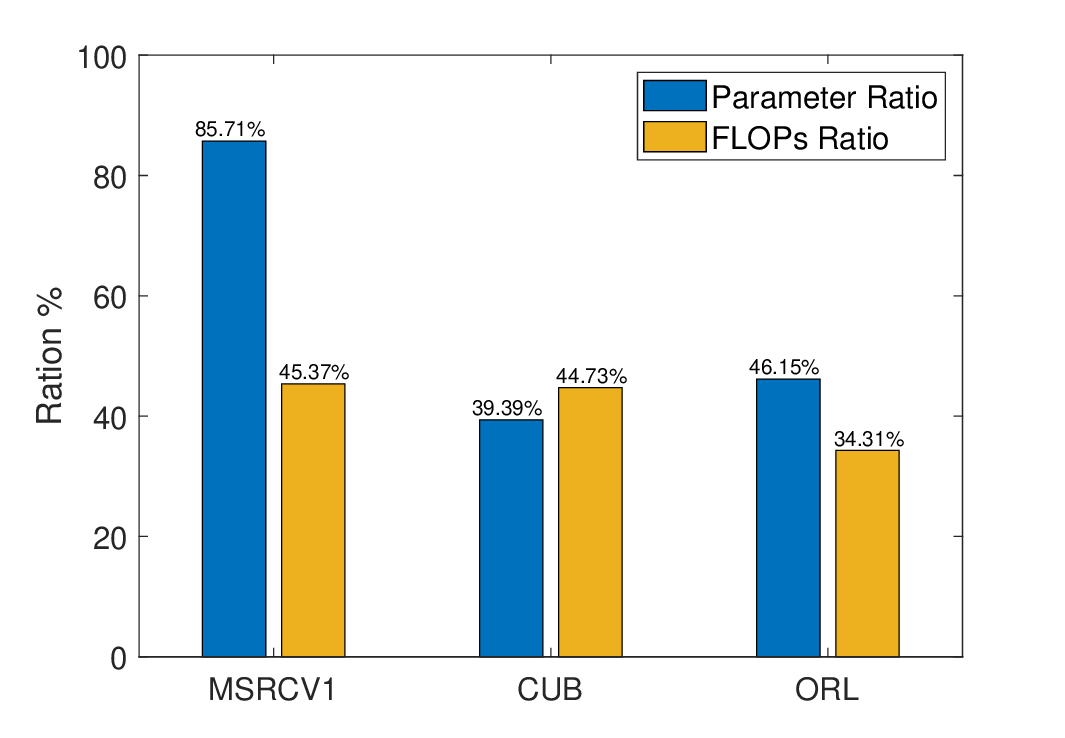}
\caption{Efficiency comparison between UMDL and AdaMuS. The chart displays the ratios of parameters and FLOPs required by AdaMuS relative to UMDL (blue and orange bars, respectively).}
\label{para-flops}
\end{figure}

\subsubsection{Evaluation Metrics}
\label{sec:metrics}

To assess the clustering performance, we employ three standard metrics: Accuracy (ACC), Normalized Mutual Information (NMI), and Adjusted Rand Index (ARI). In terms of the classification task, we utilize Accuracy (ACC) and F1-Score for evaluation.

\subsubsection{Implementation Details}
We execute the AdaMuS framework via a four-step procedure. The proposed model is developed using PyTorch 2.2.2 within a Windows 11 environment.
Step 1: Learning the Comprehensive Representation.
The multi-view comprehensive representation $\boldsymbol{z}_n$ is learned via the view-specific encoders and fusion mechanism detailed in Eq.~\eqref{eq:5} and Eq.~\eqref{eq:8}.

Step 2: Constructing Pseudo-Labels.
The consensus similarity matrix $\boldsymbol{S}$ is established by solving Eq.~\eqref{eq:10} using the Adam optimizer with a learning rate of 0.001. The pseudo-labels $\{l_{ij}\}$ are constructed from the similarity matrix $\boldsymbol{S}$.

Step 3: Training the AdaMuS Model.
The AdaMuS model is trained by optimizing the main objective function from Eq.~\eqref{eq:13}. This training stage also utilizes the Adam optimizer with a 0.001 learning rate.

Step 4: Pruning and Fine-tuning.
The pruning rate for each encoder layer is calculated according to Eq.~\eqref{eq:15}. Then we adopt One-shot pruning strategy to remove neurons and fine-tune the AdaMuS model


\subsection{Main Results}
\label{sec:main_results}

We start by reporting the quantitative performance of AdaMuS on the fundamental tasks of clustering and classification.

\subsubsection{Clustering Performance}
\label{sec:clustering_results}

Table~\ref{pcoc_journal_style_final} presents the clustering performance of AdaMuS against twelve baseline methods across six distinct multi-view datasets. The experimental results reveal several key observations:

\noindent (1) The \textbf{Concat} strategy yields sub-optimal results on the CUB benchmark, which is characterized by significant imbalance between its two views. This indicates that the naive concatenation of dimensionally disparate features often leads to the suppression of critical information embedded in the lower-dimensional view.

\noindent (2) We further observe that methods based on graph structure learning (e.g., \textbf{GUMRL}~\cite{zheng2022graph},
\textbf{SURER}~\cite{wang2024surer}) generally outperform foundational joint representation learning models (e.g., \textbf{DCCAE}~\cite{wang2015deep}) and decision-level fusion methods (e.g., \textbf{STCMC-UR}~\cite{hu2025self}). This is likely because, compared to simple correlation-based fusion or late-stage fusion, proactive view alignment strategies can more effectively integrate cross-view information. However, the performance of these methods degrades markedly on datasets with high dimensional imbalance between views, like CUB. This may suggest that their existing alignment strategies are still insufficient under unbalanced multi-view settings.

\noindent (3) As shown in Table~\ref{pcoc_journal_style_final}, AdaMuS achieves state-of-the-art performance across all tested datasets, demonstrating the effectiveness of our proposed method. In particular, when compared to \textbf{UMDL}~\cite{xu2023unbalanced}, which is a SOTA model  designed to tackle view imbalance, AdaMuS shows significant advantages. UMDL primarily addresses the overfitting issue in low-dimensional views through dictionary learning but fails to manage potential parameter redundancy in high-dimensional views. In contrast, the adaptive pruning method proposed in AdaMuS operates on the encoders of all views, dynamically removing redundant parameters based on data characteristics. This allows for a more refined treatment of each view within an end-to-end learning framework.
\begin{figure}[t]
\centering
\scalebox{0.9}{
\begin{minipage}{\linewidth} 
\centering 

\subfloat[1024D image feature view]{\label{fig:cub_a}\includegraphics[width=1.6in]{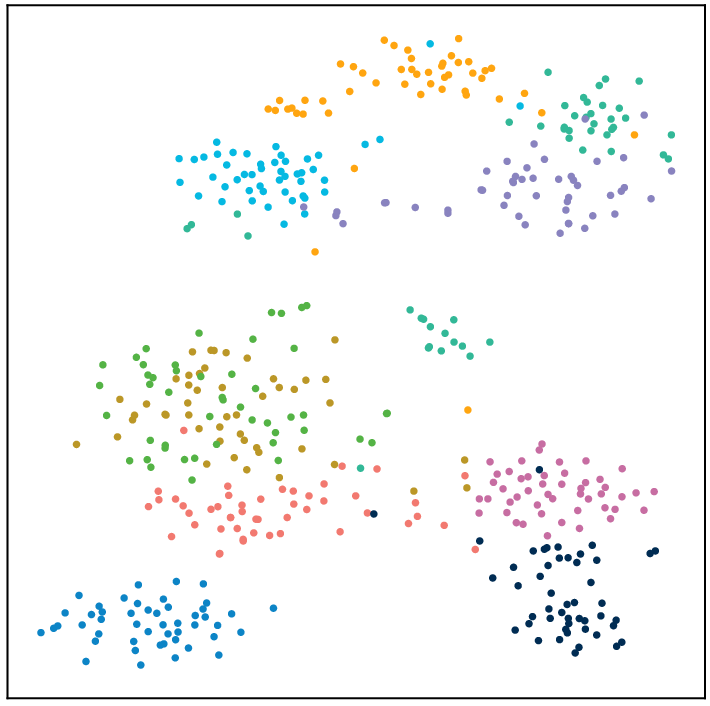}}
\hfill 
\subfloat[30D text feature view]{\label{fig:cub_b}\includegraphics[width=1.6in]{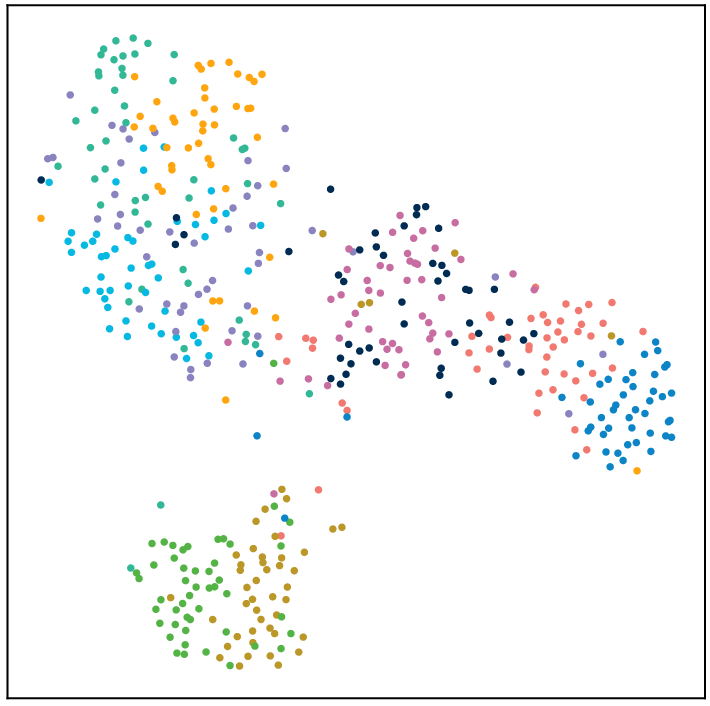}}
\vfill 
\subfloat[AdaMuS-RS]{\label{fig:cub_c}\includegraphics[width=1.6in]{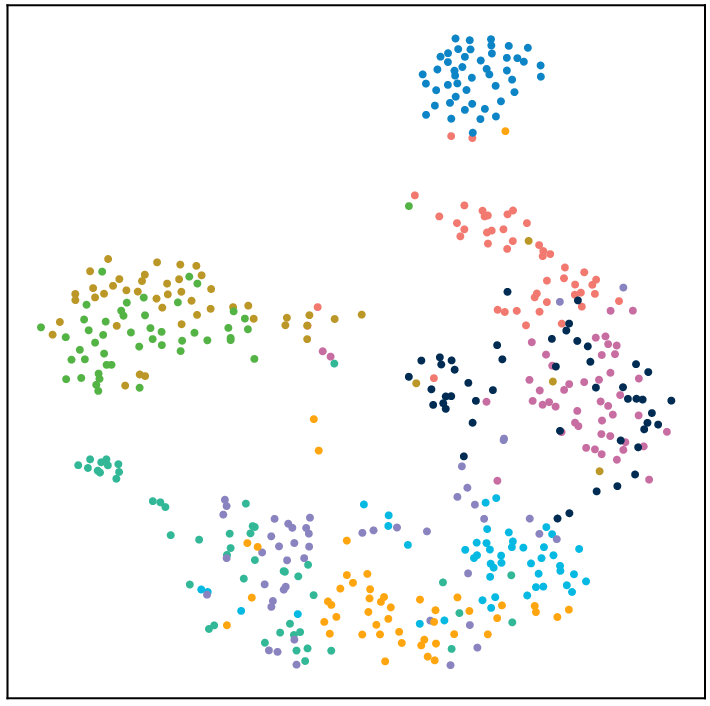}}
\hfill 
\subfloat[AdaMuS]{\label{fig:cub_d}\includegraphics[width=1.6in]{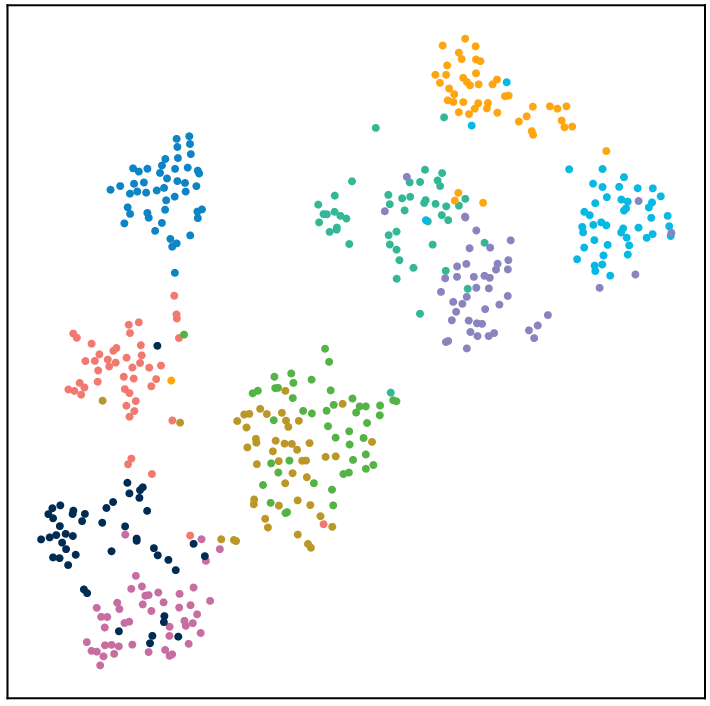}}

\end{minipage} 
}

\caption{t-SNE visualization of original data and comprehensive representations on CUB dataset. (a) and (b) are original features, respectively. (c) is the result of AdaMuS-RS, which removes the MSBN layer. (d) is the result of AdaMuS.}
\label{fig:t-sne} 
\end{figure}

\begin{table}[tb]
    \centering
    \caption{Ablation study. $\mathcal{L}_{G}$ is the foundational graph-based contrastive learning module. $\mathcal{L}_{S}$ denotes the sparsity fusion constraint, and $\mathcal{L}_{P}$ represents the PNA pruning module. Best results are in \textbf{bold}.}
    \label{tab:ablation_study_single_col}
    
    \footnotesize 
    \setlength{\tabcolsep}{5pt} 

    \begin{tabular}{c|ccc|ccc}
        \toprule
        \multirow{2}{*}{\textbf{Datasets}} & \multicolumn{3}{c|}{\textbf{Method}} & \multicolumn{3}{c}{\textbf{Metrics}} \\ 
        \cmidrule(lr){2-4} \cmidrule(lr){5-7}
        & $\mathcal{L}_{G}$ & $\mathcal{L}_{S}$ & $\mathcal{L}_{P}$ & \textbf{ACC (\%)} & \textbf{NMI (\%)} & \textbf{ARI (\%)} \\
        \midrule
        \multirow{4}{*}{\textbf{MSRCV1}}     
            & \cmark & - & - & 87.67$\pm$1.59 & 80.72$\pm$1.84 & 76.24$\pm$1.80 \\
            & \cmark & \cmark & - & 88.71$\pm$1.47 & 81.19$\pm$1.33 & 76.66$\pm$1.72 \\
            & \cmark & - & \cmark & 92.86$\pm$0.92 & 88.28$\pm$1.05 & 83.21$\pm$1.13 \\
            & \cmark & \cmark & \cmark & \textbf{95.05$\pm$0.77} & \textbf{93.07$\pm$0.76} & \textbf{91.75$\pm$0.60} \\
        \midrule           
        \multirow{4}{*}{\textbf{ORL}}     
            & \cmark & - & - & 70.27$\pm$2.34 & 84.21$\pm$1.18 & 66.04$\pm$1.83 \\
            & \cmark & \cmark & - & 72.23$\pm$1.80 & 85.68$\pm$0.76 & 66.39$\pm$1.71 \\
            & \cmark & - & \cmark & 76.89$\pm$1.73 & 87.47$\pm$0.55 & 70.20$\pm$1.43 \\
            & \cmark & \cmark & \cmark & \textbf{80.98$\pm$1.55} & \textbf{87.65$\pm$0.82} & \textbf{74.95$\pm$0.40} \\
        \midrule
        \multirow{4}{*}{\textbf{CUB}}   
            & \cmark & - & - & 71.13$\pm$1.93 & 70.55$\pm$1.48 & 69.00$\pm$1.44 \\
            & \cmark & \cmark & - & 73.02$\pm$1.66 & 72.38$\pm$1.68 & 71.89$\pm$1.87 \\
            & \cmark & - & \cmark & 76.44$\pm$2.18 & 74.07$\pm$1.33 & 71.67$\pm$1.41 \\
            & \cmark & \cmark & \cmark & \textbf{81.65$\pm$1.74} & \textbf{77.21$\pm$0.95} & \textbf{73.75$\pm$0.65} \\
        \bottomrule
    \end{tabular}
\end{table}

\subsubsection{Classification Results}
\label{sec:classification_results}

We additionally report classification results of the proposed framework in Table~\ref{classify_journal_style_updated}.

On datasets with relatively low view imbalance, such as 100Leaves ($\Lambda = 2.14$) and Mfeat ($\Lambda = 2.22$), AdaMuS surpasses other strong baseline methods. However, when evaluated on highly unbalanced datasets like UCI ($\Lambda = 2.88$) and CUB ($\Lambda = 3.22$), the performance of many competing methods drops significantly. In contrast, AdaMuS maintains its superior performance. This highlights the effectiveness of AdaMuS in learning from both balanced and unbalanced multi-view data.


\subsection{Quantitative Analysis of Pruning Efficacy}
\label{sec:pruning_analysis}

Figure~\ref{para-flops} quantitatively demonstrates the efficacy of our proposed pruning strategy by comparing AdaMuS and UMDL~\cite{xu2023unbalanced} in terms of model complexity (number of parameters and FLOPs). To ensure a fair evaluation of the pruning effect, the encoder architectures for both models were identical before the training process began.

As indicated by the ratios of parameters and FLOPs for AdaMuS relative to UMDL, the results reveal significant redundancy in the UMDL architecture. In sharp contrast, AdaMuS significantly reduces model size while lowering computational demands. For instance, when running on the CUB dataset, AdaMuS utilizes only 39.39\% of the parameters and 44.73\% of the FLOPs of UMDL.

Crucially, this significant optimization does not come at the cost of representational capability. As confirmed by the superior clustering and classification results in Table~II and Table~III, the pruned AdaMuS architecture consistently outperforms UMDL.
\begin{figure*}[t]   
    \centering
    \scalebox{0.8}{ 
    \begin{minipage}{0.25\linewidth}
         \vspace{3pt}
         \centerline{\includegraphics[width=\textwidth]{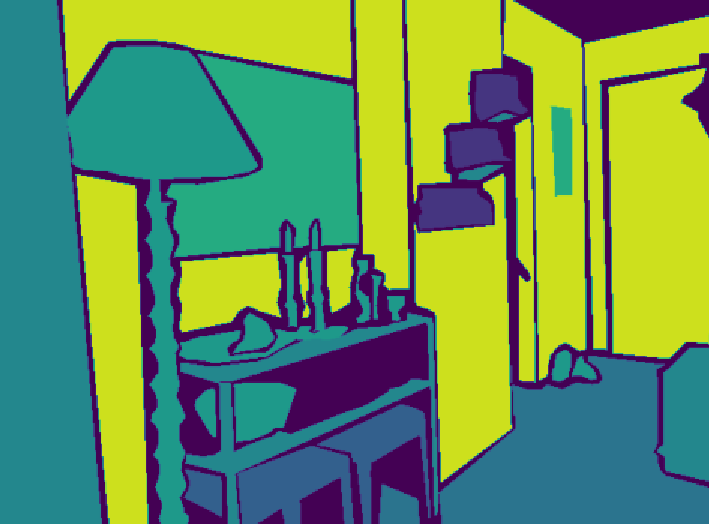}}
         \vspace{3pt}
         \centerline{\includegraphics[width=\textwidth]{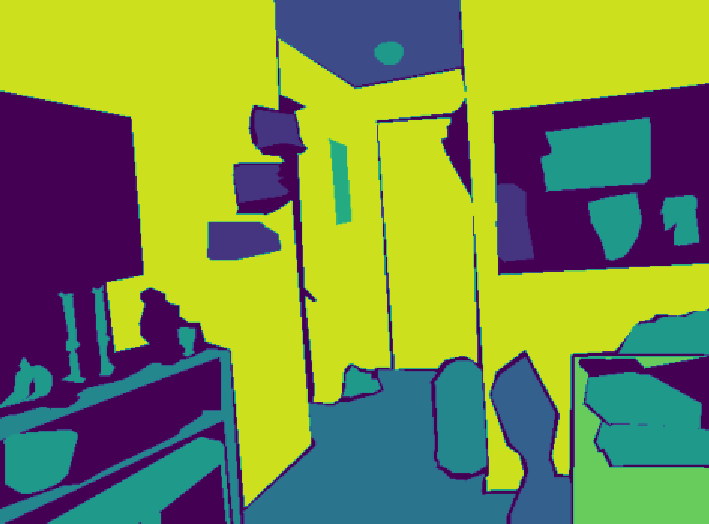}}
         \vspace{3pt}
         \centerline{\includegraphics[width=\textwidth]{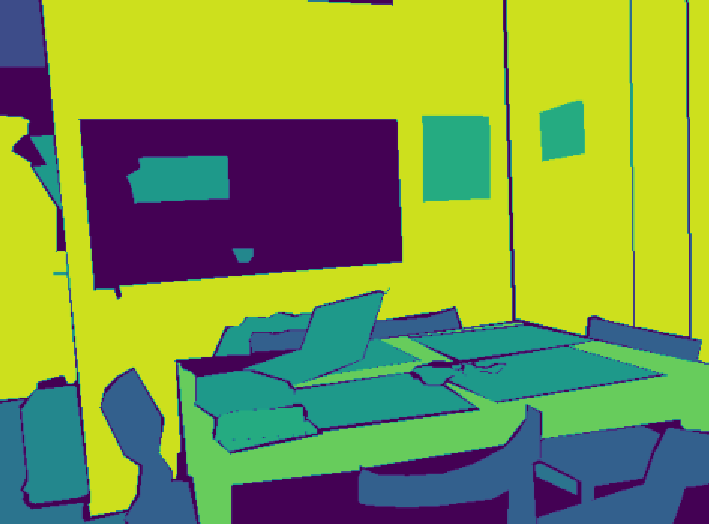}}
         \vspace{3pt}
         \centerline{Ground truth}
    \end{minipage}
    \begin{minipage}{0.25\linewidth}
         \vspace{3pt}
         \centerline{\includegraphics[width=\textwidth]{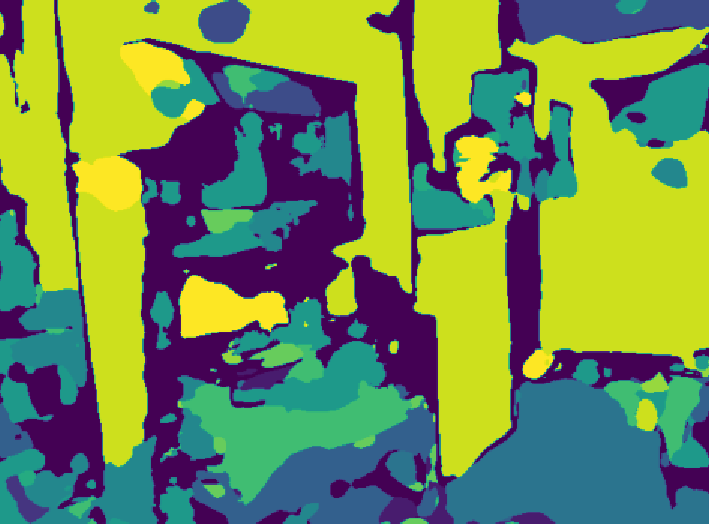}}
         \vspace{3pt}
         \centerline{\includegraphics[width=\textwidth]{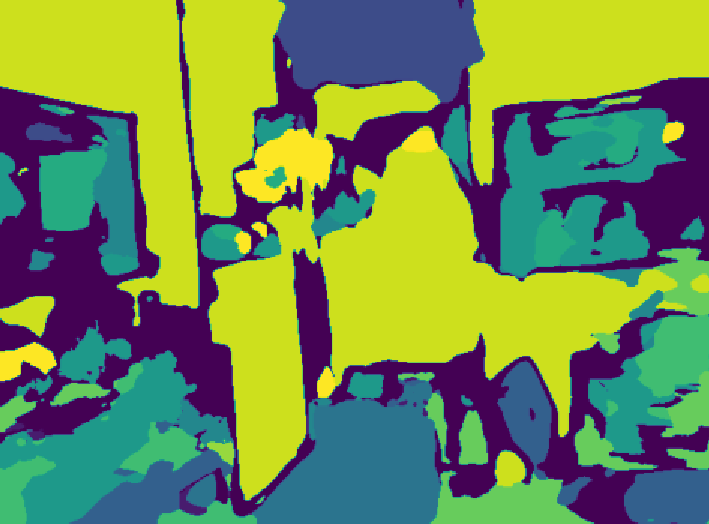}}
         \vspace{3pt}
         \centerline{\includegraphics[width=\textwidth]{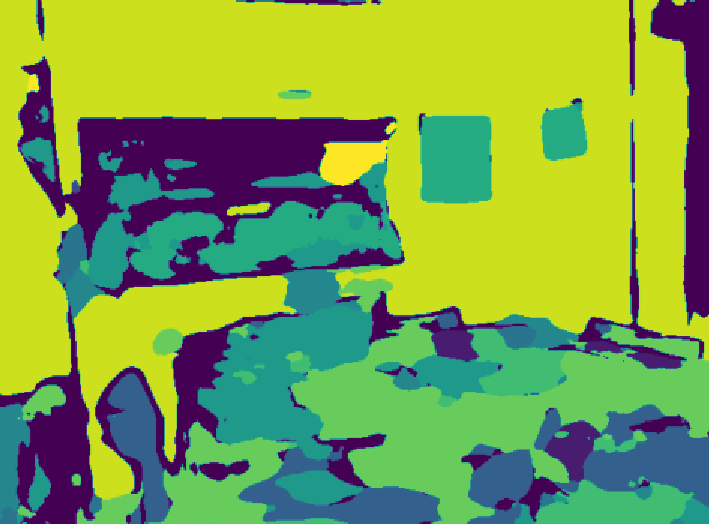}}
         \vspace{3pt}
         \centerline{AdaMuS-SEG (mIoU $22.1$)}
    \end{minipage} 
    \begin{minipage}{0.25\linewidth}
         \vspace{3pt}
         \centerline{\includegraphics[width=\textwidth]{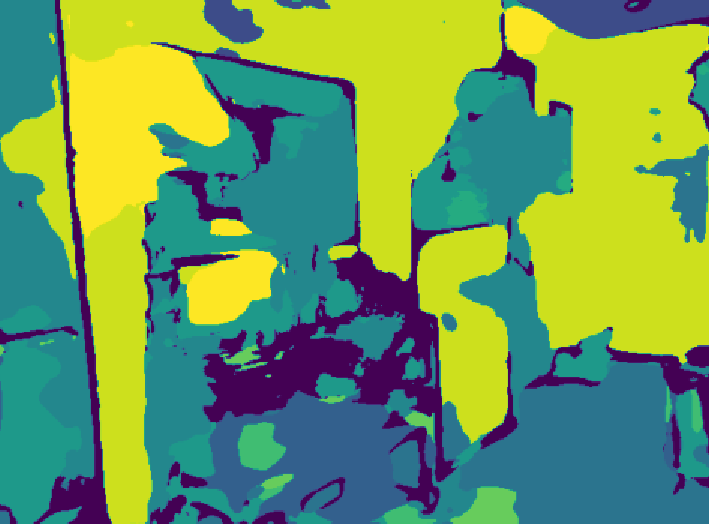}}
         \vspace{3pt}
         \centerline{\includegraphics[width=\textwidth]{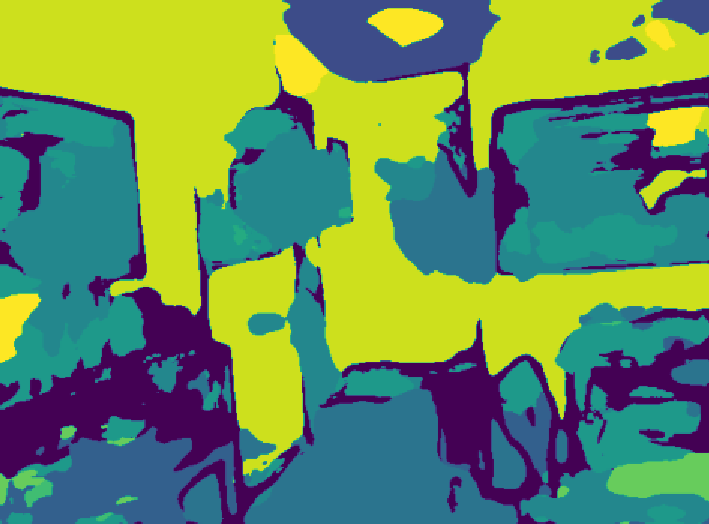}}
         \vspace{3pt}
         \centerline{\includegraphics[width=\textwidth]{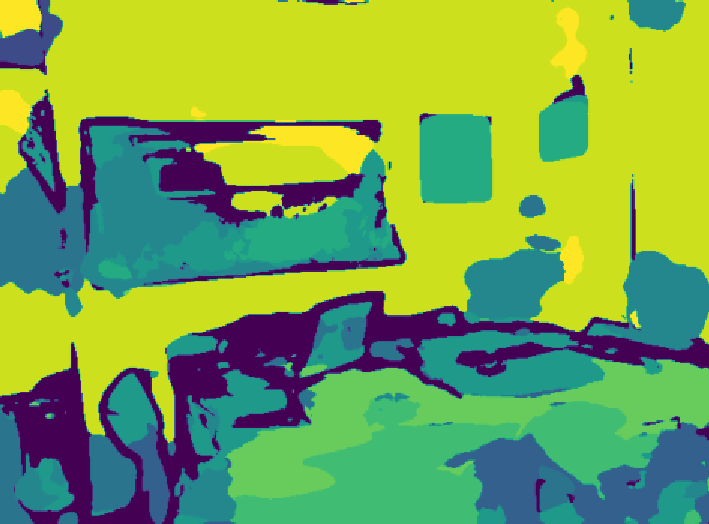}}
         \vspace{3pt}
         \centerline{UNET (mIoU $20.0$)}
    \end{minipage}
    } 
	\caption{Segmentation results of our AdaMuS-SEG and baseline UNET on the NYUv2 dataset.}
	\label{fig:segmentation}
\end{figure*}

\begin{figure*}[t]
\centering
\subfloat[Sparsity controller $\lambda_1$]{\label{fig:hyper_lambda1}\includegraphics[height=1.8in]{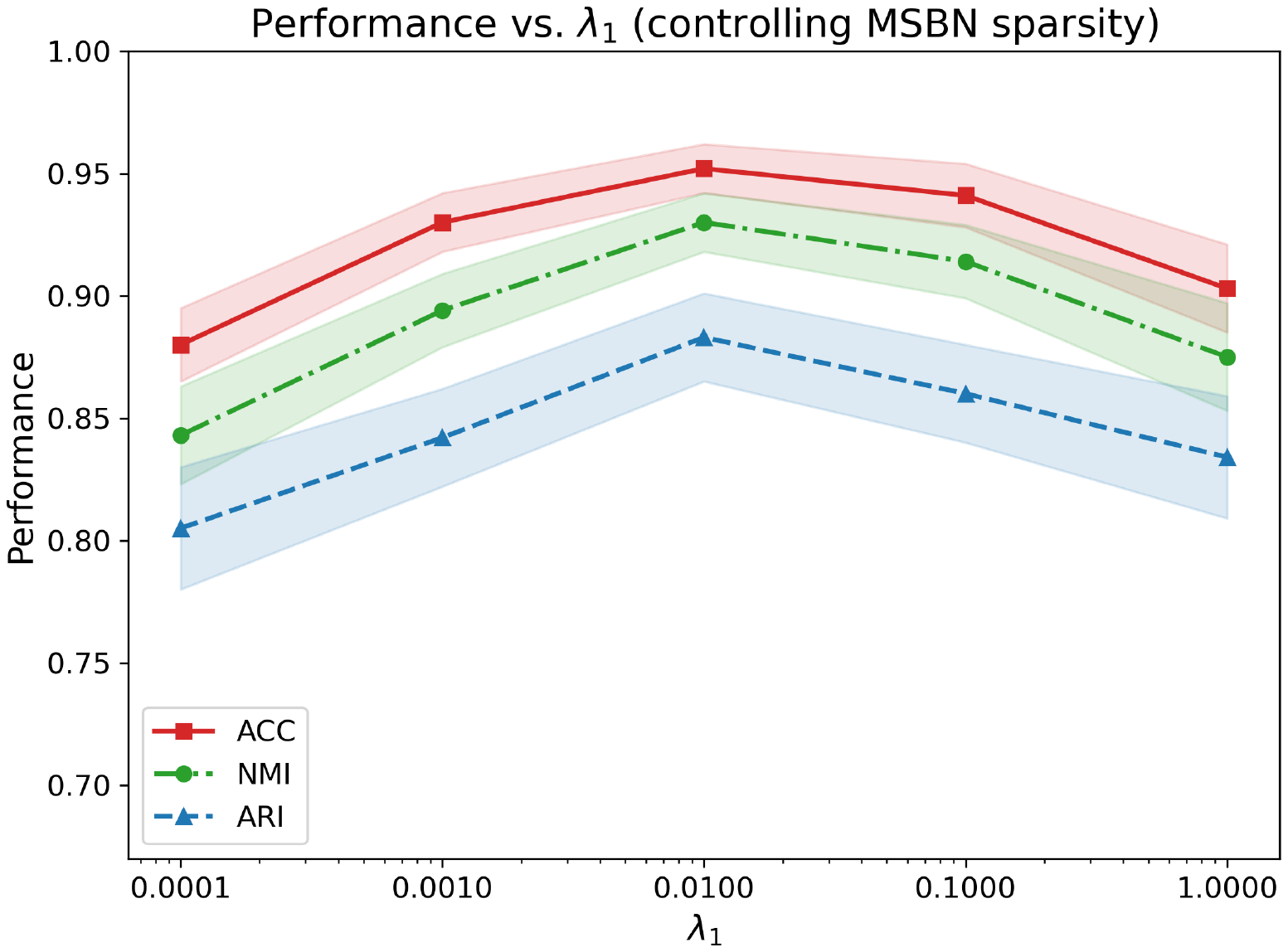}}
\subfloat[Number of neighbors $K$]{\label{fig:hyper_k}\includegraphics[height=1.8in]{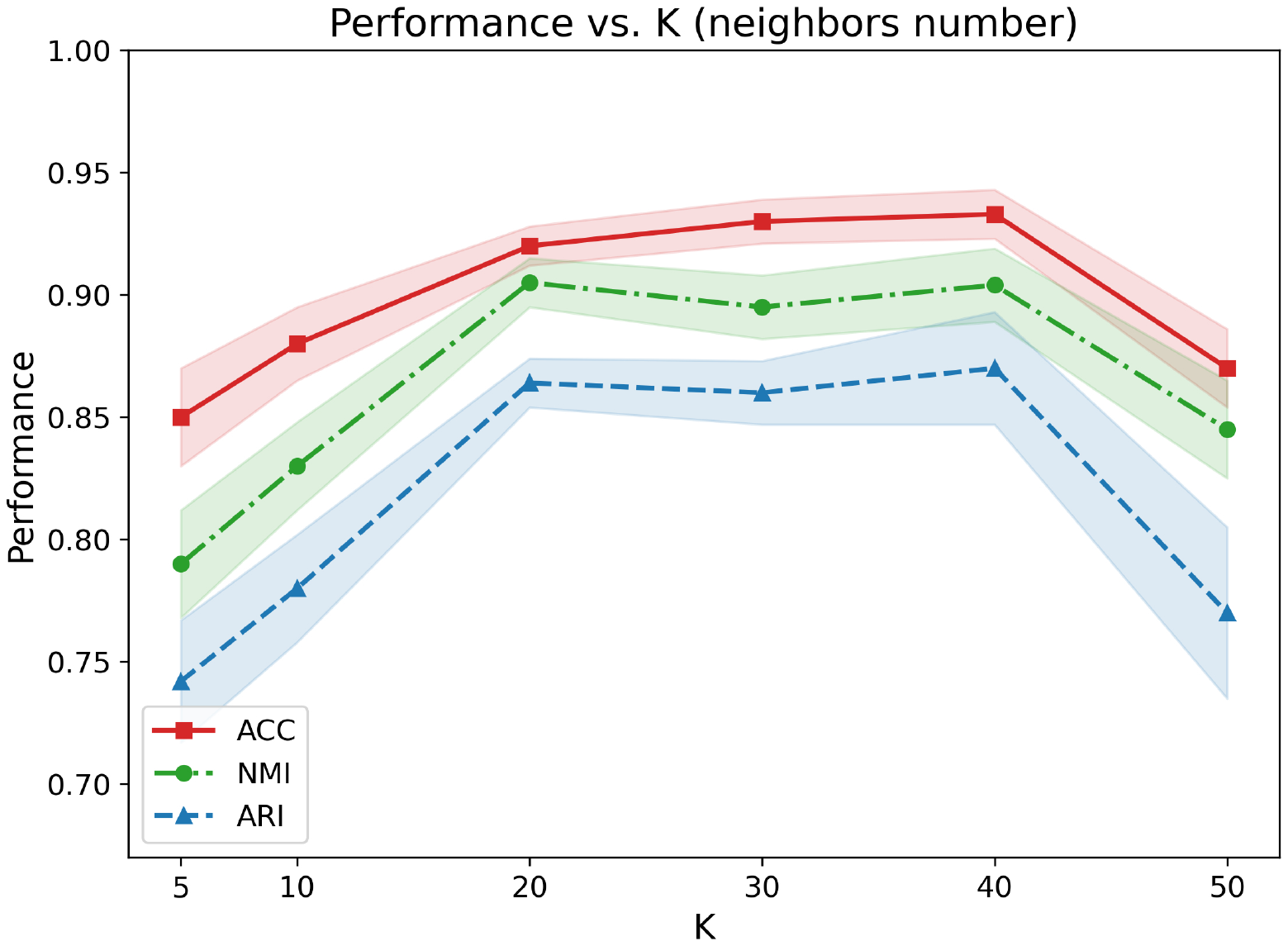}}
\subfloat[Contrastive loss margin]{\label{fig:hyper_margin}\includegraphics[height=1.8in]{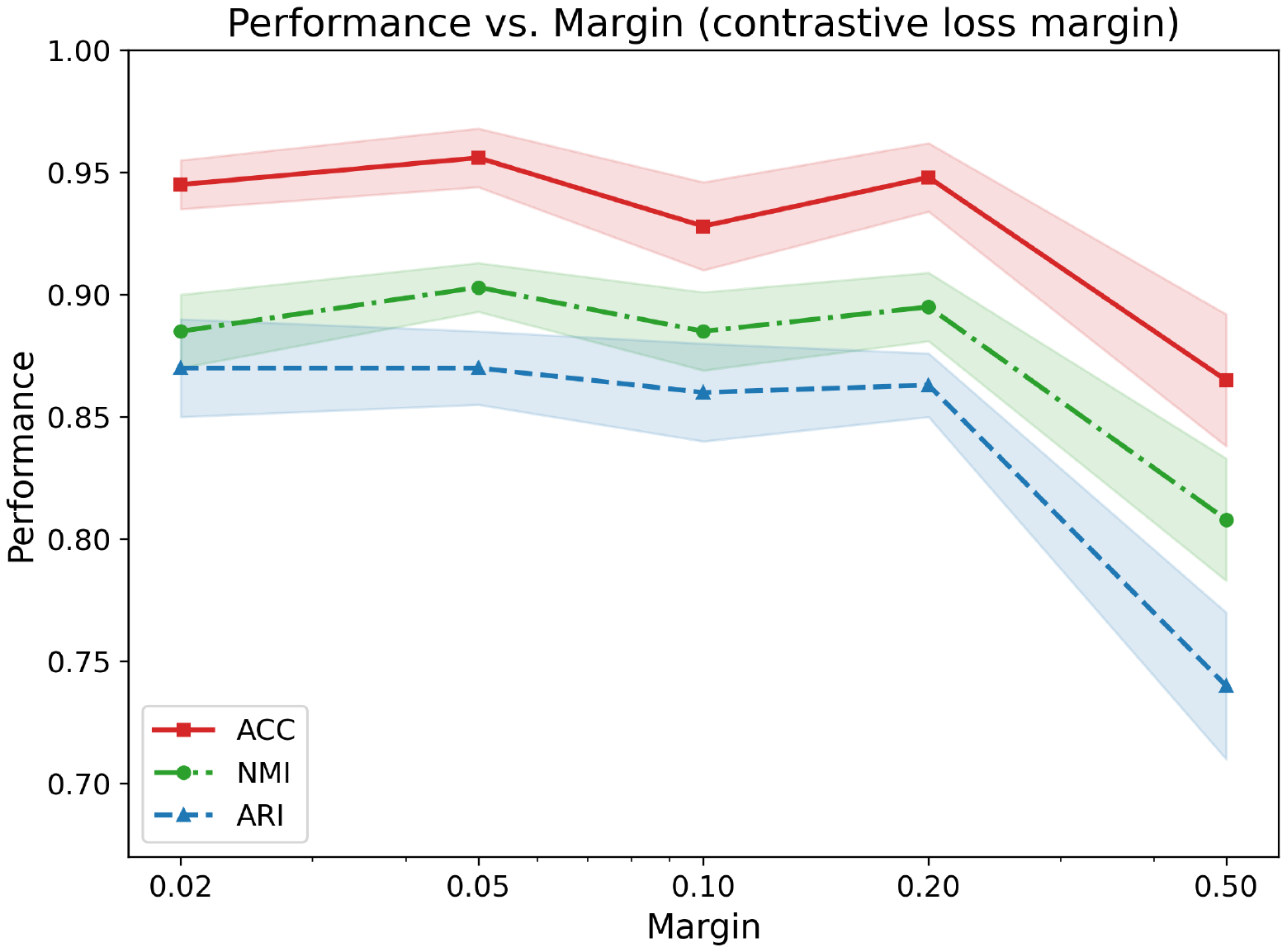}}
\caption{Hyper-parameter analysis on the MSRCV1 dataset.}
\label{fig:hyperparams}
\end{figure*}

\subsection{ Efficacy of PNA in Mitigating Overfitting }
\label{sec:pna_overfitting}

In multi-view tasks with severe dimensional imbalance between views, Multi-view Aligned Representation Learning (MARL) is an effective strategy for capturing deep inter-view interactions. 
This strategy employs DNN encoders to project multi-view features onto a shared representation with a unified dimensionality.
Such an alignment process prevents the valuable information in low-dimensional views from being overwhelmed by high-dimensional features.
However, this process introduces a large number of additional parameters (e.g. mapping a 47-dimensional feature to 200 dimensions requires 9,600 parameters), a significant portion of which may be redundant. This not only increases computational overhead but also makes the model prone to overfitting. To verify whether our proposed PNA adaptive pruning strategy can effectively mitigate this issue, we designed a dedicated single-view classification experiment.

We selected a low-dimensional view (47-dim) from the UCI dataset for a single-view classification experiment. Considering the highest dimension in the dataset is 240, we expanded the 47-dimensional features to 200 to reduce the disparity between low- and high-dimensional views. We constructed two identical classification networks (structure: 47$\rightarrow$200$\rightarrow$64$\rightarrow$10) for comparison: one was a standard DNN model trained with cross-entropy loss; the other model, based on the exact same network architecture and loss function, additionally applied our PNA adaptive pruning strategy exclusively to the up-sampling layer (47$\rightarrow$200) to specifically eliminate the redundant parameters introduced during dimensionality expansion.

The learning curves, shown in Figure~8, reveal that while the training error for both models steadily decreased, the standard DNN exhibited clear signs of overfitting as its validation error first decreased and then rebounded. In contrast, the validation error of the model equipped with the PNA adaptive pruning strategy remained stable and consistently decreased, indicating a more stable training process. This result demonstrates that  AdaMuS's adaptive pruning module can effectively reduce redundant parameters, thereby decreasing model complexity and mitigating the risk of overfitting.

\subsection{Ablation Study}
\label{sec:analysis_ablation}
We analyze the role of individual components in AdaMuS using MSRCV1, ORL, and CUB, with results reported in Table~\ref{tab:ablation_study_single_col}.
The analysis starts from a base model equipped only with the graph-guided contrastive learning loss ($\mathcal{L}_G$), and we progressively introduced the sparse fusion constraint ($\mathcal{L}_S$) and the PNA pruning module ($\mathcal{L}_P$). The results show that:

\noindent (1) \textbf{$\mathcal{L}_G$}: The baseline model already achieves competitive results, validating the effectiveness of graph-guided contrastive learning in capturing cross-view consistency.

\noindent (2) \textbf{$\mathcal{L}_G + \mathcal{L}_S$}: After introducing the sparse fusion constraint, performance improves across all datasets, which indicates that optimizing the combination of view information at the feature level has a positive effect.

\noindent (3) \textbf{$\mathcal{L}_G + \mathcal{L}_P$}: The addition of the PNA pruning module leads to a significant performance boost, demonstrating that reducing redundancy helps to enhance generalization ability.

\noindent (4) \textbf{$\mathcal{L}_G + \mathcal{L}_S + \mathcal{L}_P$ (Full Model)}: The combination of both components achieves the highest performance, illustrating that pruning and sparse fusion are complementary in optimizing representation quality and information integration.
\subsection{Qualitative Visualization of Learned Representations}
\label{sec:tsne_visualization}

We use t-SNE to visualize representations learned from the original features, AdaMuS-RS, and AdaMuS on the CUB dataset (Figure~\ref{fig:t-sne}). Here, AdaMuS-RS denotes a variant without the sparse fusion module. From the visualization, we observe that:

\noindent (1) The original low-dimensional text view exhibits a degree of separability, demonstrating that it provides complementary class information to the high-dimensional image view.

\noindent (2) The comprehensive representation learned by AdaMuS forms clearer and more well-separated clusters than its ablation variant, AdaMuS-RS. This visually validates that our sparse fusion mechanism effectively captures the complementarity of multi-view data.

\subsection{Segmentation Task Performance }
\label{sec:segmentation_analysis}

AdaMuS-SEG is first pre-trained by optimizing Eq.~\ref{eq:13} on the encoders for the RGB and depth views to learn a comprehensive multi-modal representation. Subsequently, the high-level semantic features extracted by a UNET are fused with this pre-trained representation. Finally, the entire network is fine-tuned using cross-entropy loss. The evaluation metric is the mean Intersection over Union (mIoU) of occupied voxels.

We also compare it against a standard UNET that performs segmentation directly on the RGB view. The results, visualized in Figure~9, show that AdaMuS-SEG achieves an mIoU of 22.1, while the UNET baseline reaches an mIoU of 20.0. Experimental evidence demonstrates the applicability and effectiveness of AdaMuS for dense prediction tasks like semantic segmentation.


\begin{figure}[t]
\centering
\includegraphics[width=0.65\columnwidth]{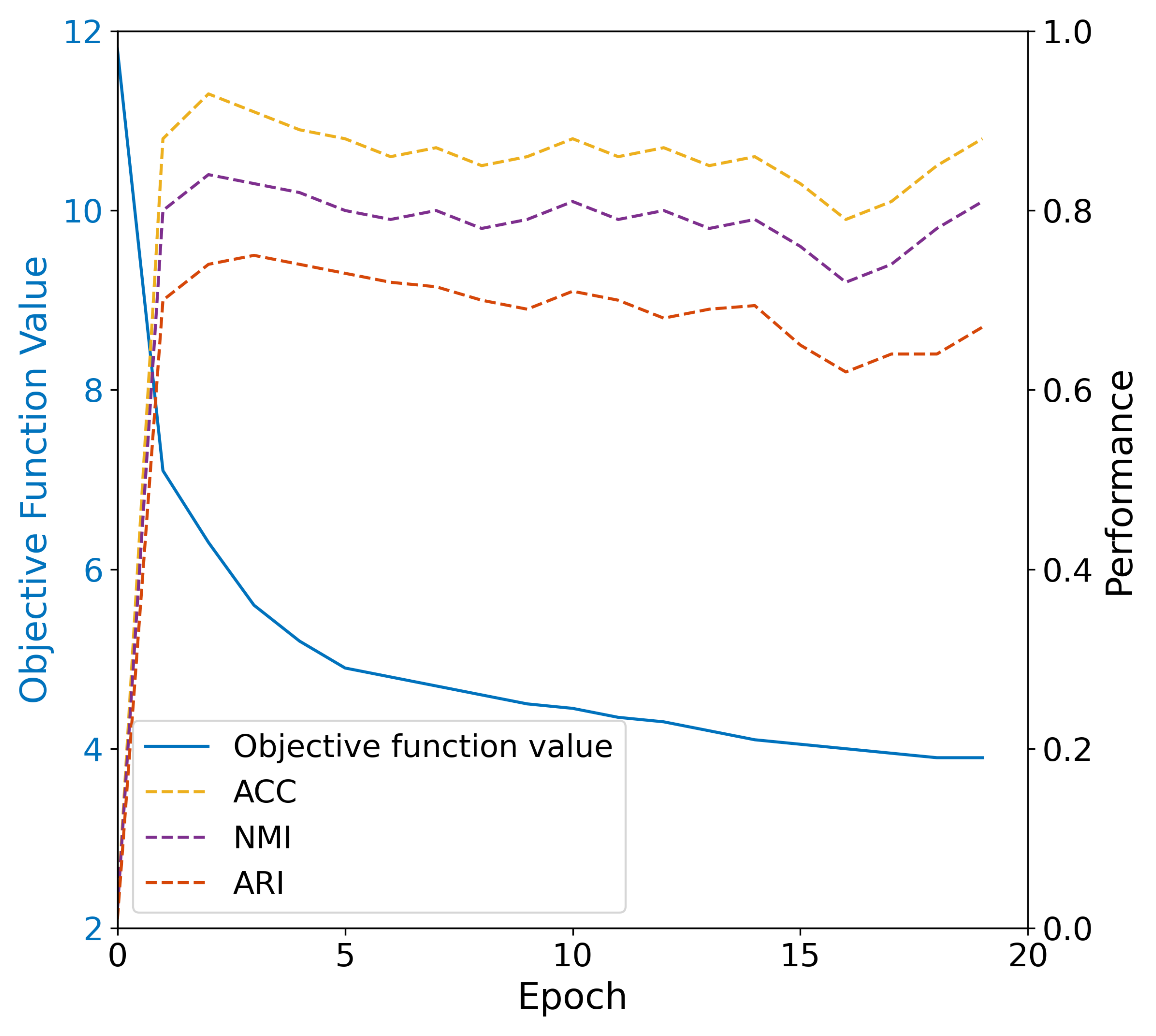}
\caption{Convergence analysis on the MSRCV1 dataset.}
\label{fig:convergence}
\end{figure}

\subsection{Hyperparameter and Convergence Analysis}
\label{sec:analysis_conv}

We analyze the impact of hyperparameters $\lambda_1$, $\lambda_2$, and K on model performance on the MSRCV1 dataset, reporting the changes in ACC, NMI, and ARI metrics (see Figure~\ref{fig:convergence}). As observed in Figure~10(b), performance first rises and then falls as $\lambda_2$ increases. This suggests that a moderate sparsity constraint is beneficial for model performance, whereas an overly strong constraint leads to performance degradation.

Figure~13 displays the curves for the loss value and the ACC, NMI, and ARI scores of AdaMuS over the training epochs on the MSRCV1 dataset. The results demonstrate that the loss value drops sharply in the initial training phase while the performance metrics rapidly improve, with the model converging within approximately 5-10 epochs.

\FloatBarrier 

\section{Conclusion}
\label{sec:conclusion}

In this work, we proposed AdaMuS, a new self-supervised framework designed to tackle the unbalanced multi-view representation learning problem. To avoid the overfitting problem, we incorporated a PNA pruning technique that effectively removed redundant encoder parameters. Furthermore, by deploying view-specific encoders with an adaptive sparse fusion paradigm, the model successfully aligned representations while filtering out cross-view redundancy to maximize complementarity. The training process was guided by self-supervision from balanced view-specific similarity graphs. The experimental results demonstrate the effectiveness of our method in handling view imbalance and supporting  downstream applications.

\bibliographystyle{./IEEEtran}
\bibliography{./IEEEabrv, reference}

\end{document}